\documentclass[sigconf]{acmart}

\usepackage{hyperref}       
\usepackage{url}           
\usepackage{booktabs}       
\usepackage{amsfonts}       
\usepackage{nicefrac}       
\usepackage{microtype}      
\usepackage{xcolor}
\usepackage{latexsym}
\usepackage{amsmath, amsthm}
\usepackage{enumitem}
\usepackage{graphicx}
\usepackage{multirow}
\usepackage{algorithm}
\usepackage{algpseudocode}
\usepackage{wrapfig,lipsum}
\usepackage{ragged2e}
\usepackage{balance}

\AtBeginDocument{%
  }

\copyrightyear{2025}
\acmYear{2025}
\setcopyright{cc}
\setcctype{by}
\acmConference[CIKM '25]{Proceedings of the 34th ACM International Conference on Information and Knowledge Management}{November 10--14, 2025}{Seoul, Republic of Korea}
\acmBooktitle{Proceedings of the 34th ACM International Conference on Information and Knowledge Management (CIKM '25), November 10--14, 2025, Seoul, Republic of Korea}\acmDOI{10.1145/3746252.3761247}
\acmISBN{979-8-4007-2040-6/2025/11}

\begin{document}

\title{EmoPerso: Enhancing Personality Detection with Self-Supervised Emotion-Aware Modelling}

\author{Lingzhi Shen}
\orcid{0000-0001-6686-0976}
\affiliation{
    \institution{School of Electronics and Computer Science, University of Southampton}
    \city{Southampton}
    \country{United Kingdom}
}
\email{l.shen@soton.ac.uk}

\author{Xiaohao Cai}
\orcid{0000-0003-0924-2834}
\affiliation{
    \institution{School of Electronics and Computer Science, University of Southampton}
    \city{Southampton}
    \country{United Kingdom}
}
\email{x.cai@soton.ac.uk}

\author{Yunfei Long}
\orcid{0000-0002-4407-578X}
\affiliation{
    \institution{School of Electronic Engineering and Computer Science, Queen Mary University of London}
    \city{London}
    \country{United Kingdom}
}
\email{yunfei.long@qmul.ac.uk}

\author{Imran Razzak}
\orcid{0000-0002-3930-6600}
\affiliation{
    \institution{Department of Computational Biology, Mohamed bin Zayed University of Artificial Intelligence}
    \city{Abu Dhabi}
    \country{United Arab Emirates}
}
\email{imran.razzak@mbzuai.ac.ae}

\author{Guanming Chen}
\orcid{0009-0002-5946-9076}
\affiliation{
    \institution{School of Electronics and Computer Science, University of Southampton}
    \city{Southampton}
    \country{United Kingdom}
}
\email{gc3n21@soton.ac.uk}

\author{Shoaib Jameel}
\orcid{0000-0001-7534-3313}
\affiliation{
    \institution{School of Electronics and Computer Science, University of Southampton}
    \city{Southampton}
    \country{United Kingdom}
}
\email{M.S.Jameel@southampton.ac.uk}

\renewcommand{\shortauthors}{Lingzhi Shen et al.}

\begin{abstract}
Personality detection from text is commonly performed by analysing users' social media posts. However, existing methods heavily rely on large-scale annotated datasets, making it challenging to obtain high-quality personality labels. Moreover, most studies treat emotion and personality as independent variables, overlooking their interactions. In this paper, we propose a novel self-supervised framework, EmoPerso, which improves personality detection through emotion-aware modelling. EmoPerso first leverages generative mechanisms for synthetic data augmentation and rich representation learning. It then extracts pseudo-labeled emotion features and jointly optimizes them with personality prediction via multi-task learning. A cross-attention module is employed to capture fine-grained interactions between personality traits and the inferred emotional representations. To further refine relational reasoning, EmoPerso adopts a self-taught strategy to enhance the model's reasoning capabilities iteratively. Extensive experiments on two benchmark datasets demonstrate that EmoPerso surpasses state-of-the-art models. The source code is available at https://github.com/slz0925/EmoPerso.
\end{abstract}



\maketitle

\section{Introduction}
Imagine browsing social media and coming across a post describing an experience on ``How to Quickly Improve Your Social Skills?''. This post resonated widely, with thousands of likes and comments \cite{cascio2023sensory}. However, its expression can actually be rewritten in different styles based on the author's personality type and emotional state. For instance, an extroverted version might emphasize active communication and group interactions, whereas an introverted version could focus more on building deep connections in small social settings \cite{tehrani2024parenting}. This phenomenon raises a fundamental question: How do personality and emotion influence the way text is expressed? Furthermore, can we leverage personality and emotion features extracted from posts to predict a user's personality type, such as the Myers-Briggs Type Indicator (MBTI)\footnote{The MBTI is a widely used personality framework that classifies individuals into 16 types based on four dichotomies: Introversion (I) vs. Extraversion (E), Sensing (S) vs. Intuition (N), Thinking (T) vs. Feeling (F), and Perceiving (P) vs. Judging (J).} \cite{butt2025interpretation}.

\begin{figure}[tbp]
    \centering
    \includegraphics[scale=0.10]{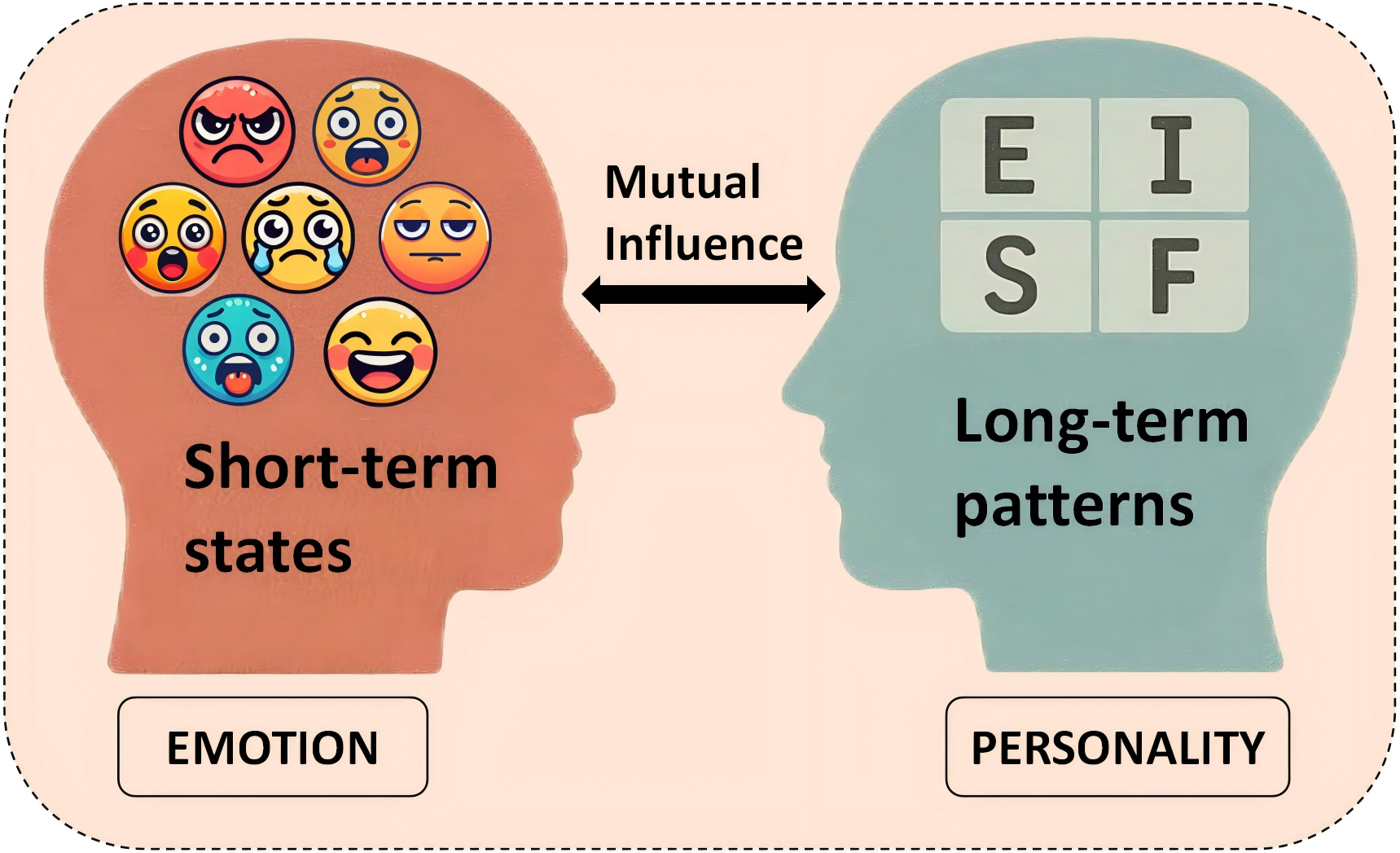} %
    \caption {An illustration of the cognitive distinction and interaction between emotional states and personality traits. 
    }
    \label{fig:placeholder}
    \vspace{-0.14in}
\end{figure}

In cognitive science, the relationship between emotion and personality has been widely studied \cite{van2024reclaiming}. Personality traits reflect an individual's long-term behavioural patterns, whereas emotions are expressions of short-term mental states \cite{william2023framework}. As illustrated in Figure~\ref{fig:placeholder}, personality and emotion operate on different cognitive timescales and interact to influence human behaviour. This perspective is also aligned with the Cognitive-Affective Personality System (CAPS) theory \citep{mischel1995cognitive}, one of the most widely cited frameworks in personality psychology, which conceptualizes personality as a stable disposition that organizes how emotional reactions vary across situations. To systematically study emotional responses, Ekman’s Basic Emotion Theory \cite{Ekman1992} provides a foundational taxonomy, positing that human emotions can be categorized into seven fundamental types: joy, anger, sadness, fear, disgust, surprise, and contempt. Building on this, studies suggest that individuals with different personality types may experience and express emotions differently when faced with the same situation \cite{yoon2024examining, teng2022understanding}. For example, extroverted individuals are more likely to exhibit joy and surprise, whereas introverted individuals may display a more reserved emotional response, such as subdued expressions of joy or a tendency toward introspective emotions like sadness or fear. 


Recent approaches have attempted to incorporate emotional information into personality prediction to improve its accuracy. For example, \citet{hu2024llm} leverage large language models (LLMs) to generate augmented textual data and interpret personality labels from raw social media posts, focusing on semantic, sentimental, and linguistic aspects. While AI-generated data can provide valuable insights, these methods often lack targeted modelling of the relationship between personality and psycholinguistic factors, making it challenging for models to capture their fine-grained interactions \cite{WangEtAl2023}. Importantly, sentiment and emotion are distinct \cite{jim2024recent}: sentiment generally reflects coarse-grained polarity (e.g., positive or negative) without capturing the nuanced psychological states conveyed by emotions. Another model, EERPD \cite{li2024eerpd}, integrates emotion regulation with emotional features using few-shot learning and chain-of-thought reasoning. However, it may require high-quality and diverse emotional data, which can be difficult to obtain in real-world scenarios. Similarly, many existing methods heavily rely on large-scale labeled datasets \cite{guerra2022datasets, wang2024multi}, yet acquiring high-quality personality-labeled data remains challenging \cite{zhu2022contrastive}, especially when input text is incomplete or noisy.

In this paper, we propose \textit{EmoPerso}, a novel self-supervised \textbf{emo}tion-\textbf{perso}nality joint learning framework, whose core idea is to infer pseudo-labeled emotion representations from personality-labeled social media posts to construct emotion-aware personality representations. To improve data diversity, EmoPerso incorporates LLM-based generative mechanisms \cite{wu2025survey}, including style-conditioned paraphrasing and contextual feature completion. Inferred emotion signals serve as auxiliary supervision and are jointly optimized with personality prediction through multi-task learning (MTL) \cite{jaradat2024multitask}, allowing the model to capture low-level sharing between emotional and personality-related features. Furthermore, the framework introduces a cross-attention module to capture fine-grained personality modulation conditioned on emotion embeddings, reinforced by an emotion-conditioned weighting mechanism that enhances the representation of psychologically salient cues. In the reasoning stage, EmoPerso integrates the Self-Taught Reasoner (STaR) \cite{zelikman2024star} to generate individualized reasoning chains, which are filtered using information gain and mutual information metrics to further strengthen the semantic coupling between emotional features and personality traits.

\noindent \textbf{Key Contributions}: Inspired by cognitive science and Basic Emotion Theory, we propose EmoPerso, a novel self-supervised emotion-personality joint learning framework, which verifies the importance of emotional features for personality detection tasks. Unlike traditional machine learning studies that treat personality prediction and emotion analysis as independent tasks while ignoring their interplay, EmoPerso leverages LLM-driven generative mechanisms, multi-task learning, cross-attention modelling, and enhanced reasoning chains, through which the interaction between pseudo-labelled emotion signals and personality traits is progressively deepened, enabling the model to gradually construct emotion-aware personality predictions. This unified strategy refines the learning process by utilizing self-generated data, introducing a flexible text augmentation paradigm that reduces reliance on external annotations. Extensive experiments demonstrate that EmoPerso outperforms state-of-the-art models on benchmark datasets.

\begin{figure*}[htbp]
    \centering
    \includegraphics[scale=0.18]{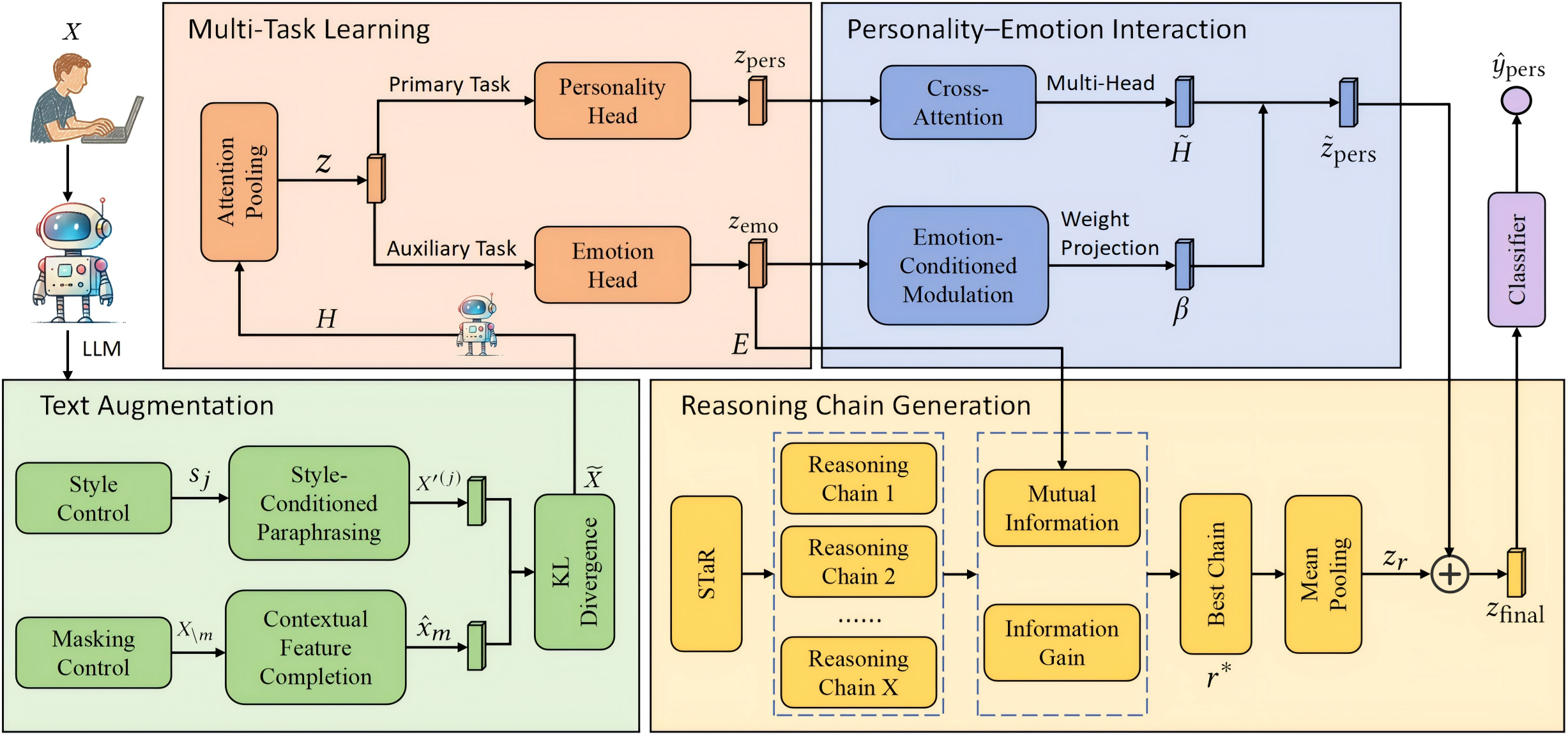} %
    \caption {Overall architecture of EmoPerso. The framework leverages LLMs for self-supervised emotion feature extraction. It integrates LLM-based generative mechanisms, MTL, cross-attention, and STaR to enhance personality-emotion interactions.}
    \label{fig:architecture}
\end{figure*}

\section{Related Work}

Previous research on personality detection spans a wide range of model architectures and learning paradigms. To provide a clear and comparable overview, we categorize existing methods into two main groups: classic deep learning methods and LLM-driven methods. 
This categorization not only reflects the methodological evolution of the field, but also highlights the shift in research focus toward cognitive reasoning.

\subsection{Classical Deep Learning Methods}
Early neural networks laid the foundation for modelling sequential and structural dependencies in user-generated content \cite{shen2025gamed}. LSTM \cite{prasanthi2021survey} and GRU \cite{shanmukha2024advancing} architectures are frequently employed to capture temporal and spatial dependencies in personality traits. Some studies have adopted hierarchical feature extraction approaches \cite{wang2022hfenet, cheng2025hierarchical}, while Transformer-based models leverage self-attention mechanisms to capture long-range dependencies, thereby improving their ability to model global relationships \cite{luo2023self}. Additionally, GNNs have been utilized to model the complex interactions between personality traits and external factors \cite{zhu2022lexical, shen2025ll4g}. However, most deep learning methods, trained in supervised settings, rely on fixed features or pretrained data \cite{zubic2024limits}, limiting adaptability and generalizability.

\subsection{LLM-Driven Methods}
Recent advances in LLMs have provided enhanced generalisation and reasoning capabilities, making them increasingly popular across a wide range of downstream tasks, including personality detection \cite{yang2023psycot}. Studies have shown \cite{bubavs2024use} that LLMs can accurately classify personality traits based on social media data and outperform traditional machine learning methods through prompt engineering and few-shot learning techniques \cite{murphy2024artificial}. Instruction-tuned models, such as GPT-4o \cite{achiam2023gpt} and Claude 3.5 Sonnet \cite{nelson2025evaluating}, have improved the reliability of personality assessment tools by generating self-evaluative texts aligned with psychological frameworks \cite{safdari2023personality}, while Llama \cite{grattafiori2024llama} has demonstrated strong performance in open-domain settings \cite{zhu2025investigating, shen2025less}. However, these methods still largely depend on shallow pattern matching rather than genuinely comprehending the logical structure of information in psychological theory reasoning tasks \cite{li2024explanation}. They also struggle to identify key information in subtle cognitive reasoning tasks \cite{aru2023mind}. 
\vspace{0.12in}

\noindent Unlike prior studies that rely on prompt engineering or externally annotated emotion data, our EmoPerso introduces a self-supervised learning framework that infers emotion representations from personality texts using pseudo-labels. These emotion features are jointly optimized with personality traits in learning process, enabling the model to internalize emotion–personality dependencies. In addition, EmoPerso incorporates LLM-based data synthesis techniques to improve generalization and employs reasoning chains to enhance its inference quality.

\section{Our Novel EmoPerso Framework}
The design of EmoPerso (Figure~\ref{fig:architecture} and Algorithm~\ref{alg:pseudocode}) introduces a novel self-supervised joint learning framework that leverages LLMs to improve personality detection through emotion-aware modelling. It enhances text generalization via LLM-based generation mechanisms, jointly optimizes emotion and personality representations through MTL, and refines their fine-grained interactions via cross-attention. Finally, STaR is employed to generate and select informative reasoning chains, further enhancing emotion-conditioned personality inference.

\begin{algorithm}[htbp]
\caption{Pseudocode of EmoPerso}
\label{alg:pseudocode}
\begin{algorithmic}[1]
\Require Dataset $\mathcal{D}$, training epochs $N$, batch size $B$, and learning rate $\eta$
\Ensure Optimized parameters $\Theta$
\State Initialize optimizer and loss weights $\lambda_{\text{MTL}}, \lambda_{\text{cross}}, \lambda_{\text{star}}$
\For{epoch in range($N$)}
    \State Load batch $(X, y_{\text{pers}})$
    \State $\widetilde{X} \gets \textsc{Augment}(X)$
    \State $z_{\text{shared}} \gets \textsc{Encode}(\widetilde{X})$
    \State $(z_{\text{pers}}, z_{\text{emo}}) \gets \textsc{Decompose}(z_{\text{shared}})$
    \State $\tilde{z}_{\text{pers}} \gets \textsc{Interact}(z_{\text{pers}}, z_{\text{emo}}, \widetilde{X})$
    \State $R \gets \textsc{GenerateChains}(X)$
    \State $\{P(r_i)\} \gets \textsc{Score}(R; z_{\text{emo}})$
    \State $r^* \gets \textsc{Select}(R; \{P(r_i)\})$
    \State $z_r \gets \textsc{Embed}(r^*)$
    \State $z_{\text{final}} \gets \textsc{Combine}(\tilde{z}_{\text{pers}}, z_r)$
    \State $\hat{y}_{\text{pers}} \gets \textsc{Infer}(z_{\text{final}})$
    \State $\mathcal{L}_{\text{MTL}} \gets \mathcal{L}_{\text{pers}} + \mathcal{L}_{\text{emo}}$
    \State $\mathcal{L}_{\text{total}} \gets \lambda_{\text{MTL}}\mathcal{L}_{\text{MTL}} + \lambda_{\text{cross}}\mathcal{L}_{\text{cross}} + \lambda_{\text{star}}\mathcal{L}_{\text{star}}$
    \State $\Theta \gets \textsc{Update}(\Theta; \nabla \mathcal{L}_{\text{total}})$
\EndFor
\end{algorithmic}
\end{algorithm}

\subsection{Text Augmentation}
\label{sec:dataaug}

We leverage large language models (LLMs) to perform self-supervised text augmentation that enhances representation diversity. Specifically, we introduce two complementary mechanisms: style-conditioned paraphrasing and contextual feature completion. Both are designed to generate semantically consistent yet informationally diverse variants of the input, facilitating better generalisation under limited labelled data.

Given an input post \( X = (x_1, x_2, \dots, x_T) \), where \( x_i \in \mathbb{V} \) and \( \mathbb{V} \) denotes the vocabulary space, we generate stylistically diverse paraphrases conditioned on a control signal \( s_j \) indicating attributes such as formality, expressiveness, or conciseness. These three styles are selected for their strong alignment with core personality dimensions, their prevalence in social media discourse, and their ability to guide generation in semantically meaningful and distinguishable ways. The LLM acts as a conditional generator, producing the \( j \)-th paraphrased variant as \( X'^{(j)} = \mathrm{LLM}(X \mid s_j) \), where \( j = 1, \dots, k \) and \( k \) is the total number of style conditions. This process relies solely on the prompt-driven capabilities of the LLM backbone.

To encourage diversity among paraphrases, we adopt sampling-based decoding strategies, specifically nucleus sampling (i.e., top-p sampling) \cite{liu2025knowledge}, instead of deterministic generation \cite{cogan2023deterministic}, which (e.g., greedy search) tends to produce repetitive and generic outputs that lack stylistic variation, limiting the model's ability to explore diverse surface realisations of personality-related content. This decoding strategy is natively supported by LLMs and is invoked during generation to promote lexical and stylistic variability.

To handle incomplete or noisy inputs, we simulate missing information by randomly masking spans in \( X \), resulting in \( X_{\backslash m} \), where \( m \subseteq \{1, \dots, T\} \) indicates masked positions. The LLM is then prompted to reconstruct the missing content based on surrounding context, predicting the masked tokens as \( \hat{x}_m = \mathrm{LLM}(X_{\backslash m}) \). Similar to paraphrasing, contextual completion also leverages the inherent infilling capability of the LLM. In practice, the masked spans are sampled at the phrase or content-word level rather than the token level to ensure syntactic and semantic plausibility.

To ensure semantic fidelity between original and augmented sequences, we impose a regularisation loss based on Kullback-Leibler (KL) divergence \cite{cui2025generalized} between their token-level output distributions. Let \( P(x_t \mid X) \) and \( P(x_t \mid \hat{X}) \) denote the predicted token distributions at position \( t \) from the LLM, conditioned on the original and augmented inputs, respectively. KL divergence is computed over aligned token positions as:
$\mathcal{L}_{\text{KL}} = \frac{1}{T} \sum_{t=1}^T D_{\text{KL}} ( P(x_t \mid X) \,\|\, P(x_t \mid \hat{X}) ).
$
This regularisation assumes access to the LLM's token-level softmax probabilities, feasible in open-source implementations.

In practice, we access token distributions via the LLM's softmax outputs at each generation step. To account for stylistic variations, we also include a supervised style classification loss \( \mathcal{L}_{\text{style}} \) using pseudo-labels corresponding to each style condition. This component is implemented using a lightweight two-layer MLP trained jointly with the main objectives. The overall augmentation loss is thus defined as:
$
\mathcal{L}_{\text{gen}} = \lambda_{\text{style}} \mathcal{L}_{\text{style}} + \lambda_{\text{KL}} \mathcal{L}_{\text{KL}},
$
where \( \lambda_{\text{style}} \) and \( \lambda_{\text{KL}} \) are balancing coefficients.

\begin{figure}[htbp]
    \centering
    \includegraphics[scale=0.170]{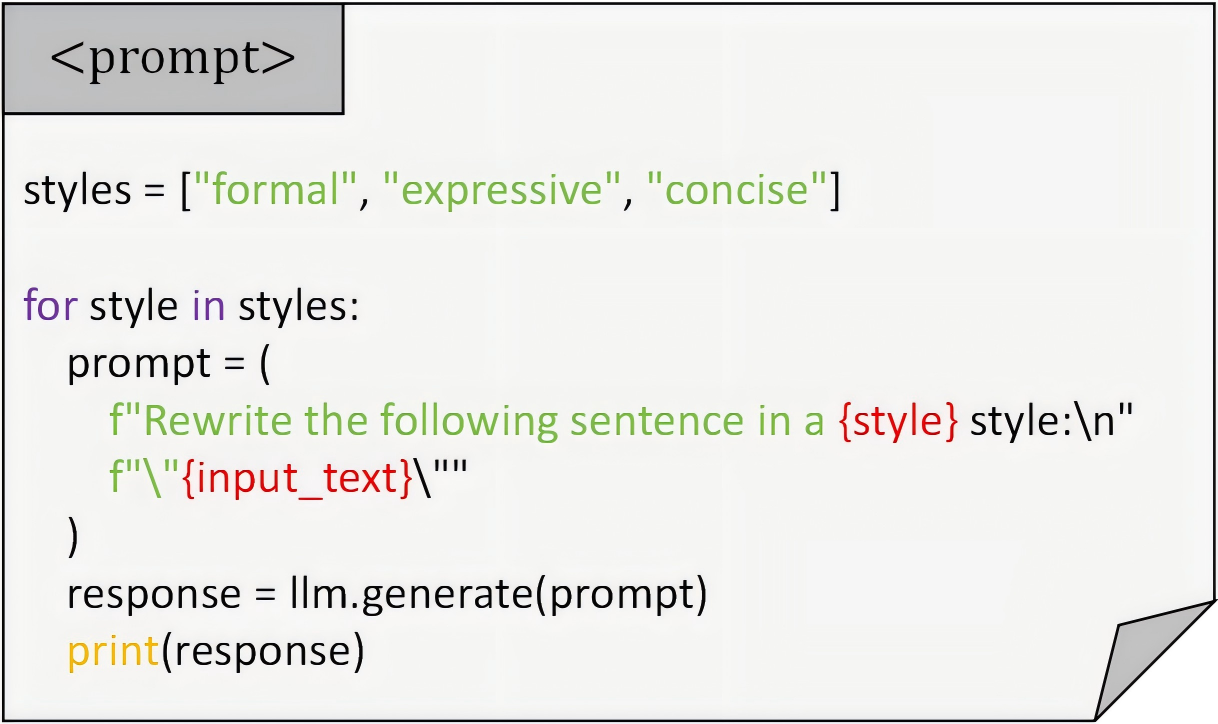}\par
    \vspace{0.1em}
    \includegraphics[scale=0.170]{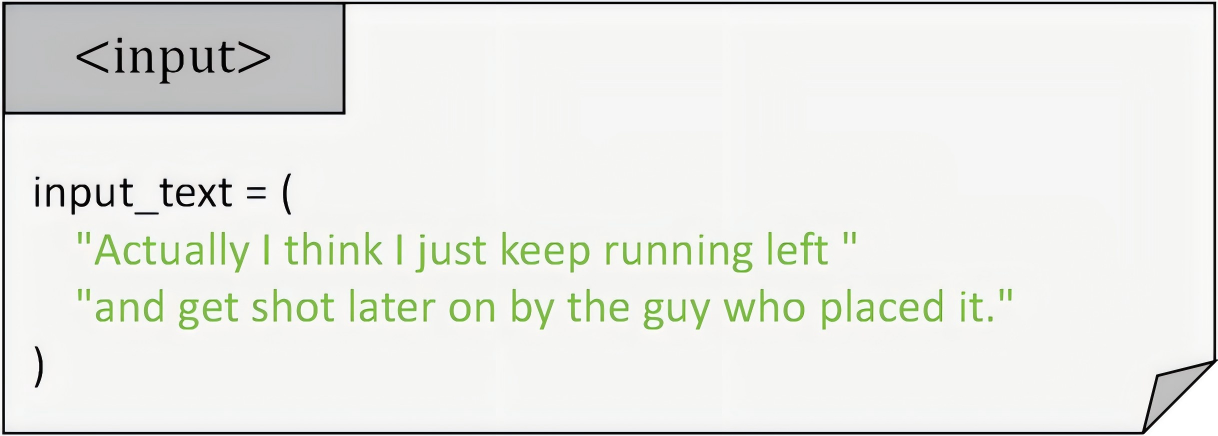}\par
    \vspace{0.1em}
    \includegraphics[scale=0.170]{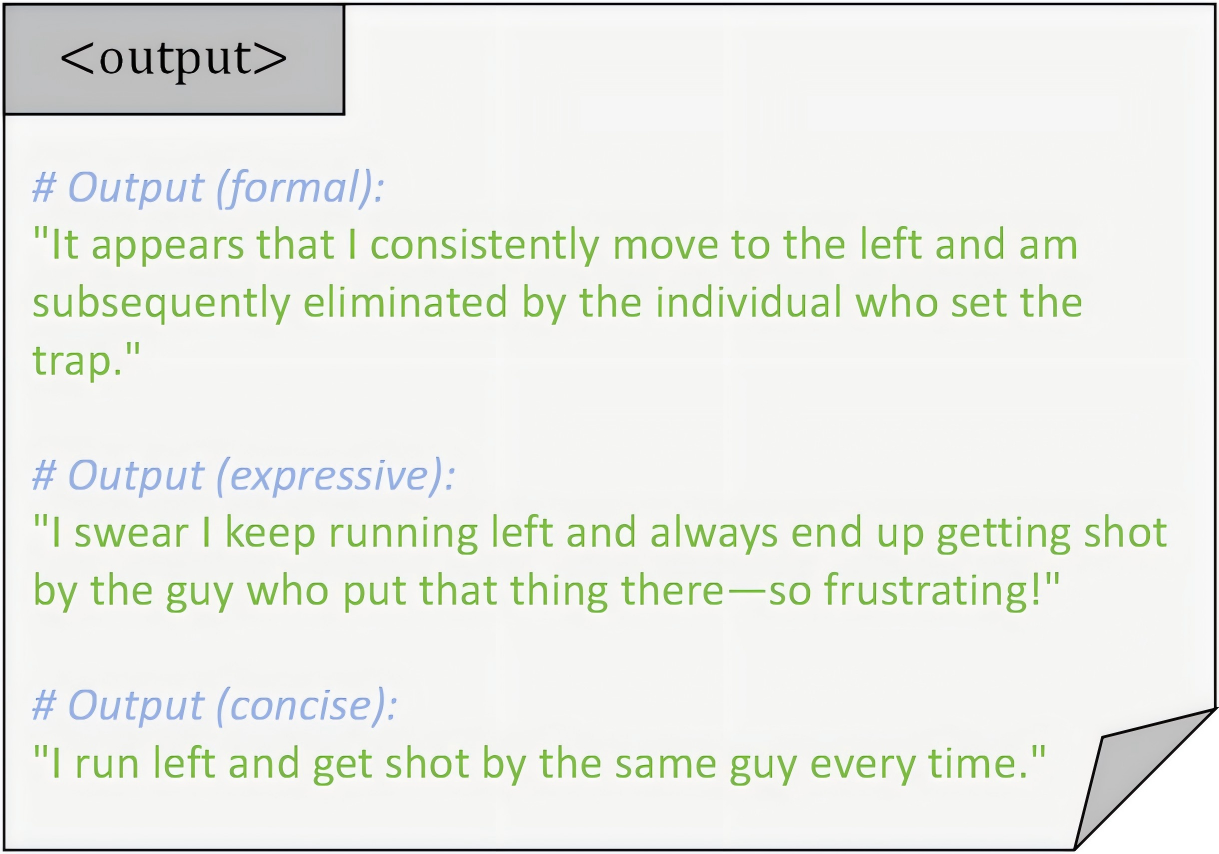}
    \caption{Code-style illustration of style-conditioned paraphrasing using a real example from the Kaggle dataset.}
    \label{fig:paraphrase}
    \vspace{-0.07in}
\end{figure}

This augmentation strategy not only mitigates data sparsity but also exposes the model to emotionally and stylistically conditioned variants, which are critical for capturing fine-grained personality signals. Example prompts and style control templates are provided in Figure~\ref{fig:paraphrase}. Given a fixed input sentence and a list of target styles, the model dynamically constructs prompts and generates stylistically diverse outputs using an LLM.

\subsection{Multi-Task Learning}

We adopt an LLM as a self-supervised feature extractor to jointly capture signals related to both personality and emotion, enabling MTL over shared latent representations. Unlike conventional approaches that treat personality and emotion as separate tasks trained on annotated datasets \cite{li2022multitask}, the core self-supervised property of our framework lies in the fact that emotion labels are not externally provided. Instead, emotion signals are inferred directly from personality-labeled data through auxiliary modelling, without any explicit supervision. The emotion head is trained using pseudo-labels derived from stylistic and emotional cues in the input, such as lexical choice, punctuation usage, and emotional valence intensity. These automatically generated signals serve as supervisory targets, allowing EmoPerso to construct emotion-aware representations in a purely self-supervised manner throughout the entire training process.

Given an augmented input post \( \widetilde{X} = (\tilde{x}_1, \tilde{x}_2, \dots, \tilde{x}_T) \), produced via style transformation and feature completion, the LLM encodes the sequence into a hidden representation \( H = (h_1, h_2, \dots, h_T) \), where \( h_i \in \mathbb{R}^d \) denotes the embedding of token \( \tilde{x}_i \).

To aggregate token-level information into a global representation, we apply an attention pooling mechanism over \( H \), i.e.,
\begin{equation}
z = \sum_{i=1}^{T} \alpha_i h_i, \quad
\alpha_i = \frac{\exp\left( v^\top \tanh(W h_i + b) \right)}{\sum_{j=1}^{T} \exp\left( v^\top \tanh(W h_j + b) \right)},
\end{equation}
where \( W \in \mathbb{R}^{d' \times d} \), \( b \in \mathbb{R}^{d'} \), and \( v \in \mathbb{R}^{d'} \) are trainable parameters. Technically, this attention pooling mechanism is implemented as a single-head feedforward scoring function using a two-layer MLP followed by softmax normalisation. This design enhances the model's ability to focus on semantically salient tokens, which is especially important for personality and emotion modelling, where relevant signals are often sparse and context-dependent.

We treat personality detection as four independent binary classification tasks, corresponding to the MBTI dimensions: Introversion/Extraversion (I/E), Sensing/Intuition (S/N), Thinking/Feeling (T/F), and Perceiving/Judging (P/J). Let \( C_p = 4 \) be the number of dimensions. Both personality and emotion tasks project the shared representation \( z \) into task-specific embeddings using lightweight two-layer MLPs, denoted as \( z_{\text{pers}} = f_{\text{pers}}(z) \) and \( z_{\text{emo}} = f_{\text{emo}}(z) \), where \( z_{\text{pers}}, z_{\text{emo}} \in \mathbb{R}^d \) represent personality-specific and emotion-specific representations, respectively. These embeddings are used to produce the corresponding classification logits for each task.

The personality prediction head outputs a 4-dimensional logit vector, and the probability for each dimension is computed using sigmoid activation as \( p_c = \sigma\left((W_{\text{pers}} z_{\text{pers}} + b_{\text{pers}})_c\right) \), where \( c = 1, \dots, C_p \), \( W_{\text{pers}} \in \mathbb{R}^{C_p \times d} \), and \( b_{\text{pers}} \in \mathbb{R}^{C_p} \) are learnable parameters. The corresponding binary cross-entropy loss is:
\begin{equation}
\mathcal{L}_{\text{pers}} = - \sum_{c=1}^{C_p} \left( y_c \log p_c + (1 - y_c) \log (1 - p_c) \right),
\end{equation}
where \( y_c \in \{0, 1\} \) is the ground-truth label for the \( c \)-th personality dimension.

Emotion prediction is modelled as a multi-label classification task over \( C_e \) emotion categories. Importantly, these emotion labels are not provided by external annotation. Instead, the model learns to predict emotion categories based on latent emotional cues that co-occur in personality-labeled text. These cues include stylistic and psycholinguistic markers such as valence-bearing adjectives (e.g., “excited,” “frustrated”), intensifiers (e.g., “really,” “extremely”), exclamation usage, emotive punctuation, and affective n-grams, which are often indicative of underlying emotional states. This enables the emotion stream to act as a self-supervised auxiliary task. The prediction is computed using sigmoid activation as \( \tilde{p}_c = \sigma\left((W_{\text{emo}} z_{\text{emo}} + b_{\text{emo}})_c\right) \), where \( c = 1, \dots, C_e \), \( W_{\text{emo}} \in \mathbb{R}^{C_e \times d} \), and \( b_{\text{emo}} \in \mathbb{R}^{C_e} \) are trainable parameters. The corresponding multi-label binary cross-entropy loss is:
\begin{equation}
\mathcal{L}_{\text{emo}} = - \sum_{c=1}^{C_e} \left( y_c \log \tilde{p}_c + (1 - y_c) \log (1 - \tilde{p}_c) \right),
\end{equation}
where \( y_c \in \{0,1\} \) denotes the pseudo-label for the \( c \)-th emotion category, selected from the inferred label set \( \hat{Y}_{\text{emo}} \) automatically constructed based on emotional cues in the input.

The overall MTL objective jointly optimises both tasks, i.e.,
\begin{equation}
\mathcal{L}_{\text{MTL}} = \lambda_{\text{pers}} \mathcal{L}_{\text{pers}} + \lambda_{\text{emo}} \mathcal{L}_{\text{emo}},
\end{equation}
where \( \lambda_{\text{pers}} \) and \( \lambda_{\text{emo}} \) are hyperparameters balancing the two tasks.

\subsection{Personality–Emotion Interaction}

Personality traits and emotional states, while related, often exhibit asymmetric and context-dependent correlations at the token level \cite{pliskin2020proposing}. For example, emotional expressions tend to occur sparsely in text but have varying influences across different personality dimensions. To realize this intuition, we design a two-stage interaction mechanism: a multi-head cross-attention module that aligns personality with token-level input, and an emotion-conditioned modulation layer that reweights token contributions based on emotional context. This design enables the model to align emotionally salient input regions with personality-relevant features, going beyond standard attention layers that treat auxiliary signals as uniformly distributed context \cite{li2023uniformer}.

Specifically, to enhance fine-grained interaction modelling between personality and emotion traits, we refine the personality-specific representation \( z_{\text{pers}} \), which is obtained from the MTL head, via a multi-head cross-attention mechanism. This step enables the model to selectively attend to relevant contextual tokens based on personality semantics, while later incorporating emotion-conditioned modulation to further refine the final representation.

Given the hidden token sequence \( H = (h_1, h_2, \dots, h_T) \in \mathbb{R}^{T \times d} \) and the personality-specific query vector \( z_{\text{pers}} \in \mathbb{R}^{d} \), we compute multi-head cross-attention by projecting \( z_{\text{pers}} \) as a query and \( H \) as keys and values. Specifically, for each head \( h \in \{1, \dots, H\} \), the query, key, and value matrices are computed as:
\begin{equation}
Q^{(h)} = W_Q^{(h)} z_{\text{pers}}, \quad K^{(h)} = W_K^{(h)} H, \quad V^{(h)} = W_V^{(h)} H,
\end{equation}
where \( Q^{(h)} \in \mathbb{R}^{1 \times d_k} \), \( K^{(h)}, V^{(h)} \in \mathbb{R}^{T \times d_k} \), and all \( W_*^{(h)} \) are head-specific trainable projection matrices. This attention mechanism is implemented using standard scaled dot-product attention, where the outputs are computed as:
\begin{equation}
A^{(h)} = \text{Softmax}\left( \frac{Q^{(h)} {K^{(h)}}^\top}{\sqrt{d_k}} \right) V^{(h)}.
\end{equation}

The resulting attended token representations \( \tilde{H} = (\tilde{h}_1, \tilde{h}_2, \dots, \tilde{h}_T) \) are obtained by concatenating outputs from all heads and projecting them via a learned output matrix \( W_O \in \mathbb{R}^{Hd_k \times d} \), i.e., each token representation is computed as \( \tilde{h}_i = \text{Concat}(A_i^{(1)}, \dots, A_i^{(H)}) W_O \). 

To further incorporate emotion-specific context, we introduce an emotion-conditioned attention modulation. Based on the emotion embedding \( z_{\text{emo}} \in \mathbb{R}^d \), predicted from pseudo-label-guided emotion supervision in the MTL stage, we compute token-level importance weights as \( \beta = \text{Softmax}(W_{\text{emo}} z_{\text{emo}}) \), where \( W_{\text{emo}} \in \mathbb{R}^{T \times d} \) is a learned projection matrix. These weights are used to aggregate the attended token representations such that the final personality vector is given by \( \tilde{z}_{\text{pers}} = \sum_{i=1}^{T} \beta_i \tilde{h}_i \). This design allows the model to selectively amplify psychologically salient features from the diverse semantic subspaces constructed in the previous cross-attention step, guided by emotional context.

To align personality- and emotion-guided features, we introduce a consistency regularisation loss based on cosine similarity, defined as \( \mathcal{L}_{\text{cross}} = 1 - \cos(\tilde{z}_{\text{pers}}, z_{\text{emo}}) \), where \( \cos(\cdot, \cdot) \) denotes the cosine similarity between the final personality representation and the emotion embedding. This consistency loss does not simply align representations geometrically, but instead guides the model to preserve emotionally discriminative features within the personality space by minimizing the distance between the emotion embedding and the personality representation.

The final training objective integrates this regularisation with the multi-task loss:
\begin{equation}
\mathcal{L}_{\text{total}} = \lambda_{\text{MTL}} \mathcal{L}_{\text{MTL}} + \lambda_{\text{cross}} \mathcal{L}_{\text{cross}},
\end{equation}
where \( \lambda_{\text{MTL}} \) and \( \lambda_{\text{cross}} \) are trade-off hyperparameters.

\subsection{Reasoning Chain Generation}
To further enhance the model's reasoning capabilities, we adopt a self-taught rationale generation strategy inspired by STaR, encouraging the model to generate its intermediate reasoning chains. We extend this process by introducing an information-theoretic selection mechanism to identify the most informative reasoning chains, particularly those capturing relational cues between personality and emotion features.

For each input post \( X \), we generate a set of candidate reasoning chains \( R = \{ r_1, r_2, \dots, r_n \} \), where each \( r_i = (s_1^{(i)}, s_2^{(i)}, \dots, s_L^{(i)}) \) consists of a sequence of intermediate reasoning steps, with \( s_j^{(i)} \) denoting the \( j \)-th step and \( L \) the chain length. We use an LLM decoder to autoregressively generate each reasoning step, employing nucleus sampling with temperature control to encourage diverse and coherent chains, i.e.,
\begin{equation}
s_j^{(i)} = \arg\max_{s} P(s \mid s_1^{(i)}, \dots, s_{j-1}^{(i)}, X),
\end{equation}
where \( P(\cdot) \) denotes the token-level generation distribution.

Since not all reasoning chains contribute equally to personality prediction, we apply a filtering mechanism based on information-theoretic criteria. The information gain (IG) \cite{sun2025information} measures how much uncertainty is reduced in personality classification given a reasoning chain
$\text{IG}(r_i) = H(Y) - H(Y \mid r_i),$
where \( H(\cdot) \) denotes the entropy of the predicted label distribution. In practice, both terms are approximated using the model’s predicted log-probabilities with and without conditioning on the reasoning chain \( r_i \).

Additionally, to capture the strength of emotion–personality coupling, we compute the mutual information (MI) \cite{wu2025interpreting} between each reasoning chain and the extracted emotion features, i.e.,
\begin{equation}
\text{MI}(r_i, E) = \sum_{y \in Y} \sum_{e \in E} P(y, e) \log \frac{P(y, e)}{P(y) P(e)},
\end{equation}
where \( E \) denotes the emotion feature set extracted via self-supervised learning, conditioned on pseudo-labels from the MTL module. Joint and marginal distributions \( P(y, e) \) are estimated over mini-batches using empirical frequency counts from predicted outputs. We define a normalised preference score over candidate chains as:
\begin{equation}
P(r_i) = \frac{\exp\left( \lambda_{\text{IG}} \, \text{IG}(r_i) + \lambda_{\text{MI}} \, \text{MI}(r_i, E) \right)}{\sum_{j=1}^{n} \exp\left( \lambda_{\text{IG}} \, \text{IG}(r_j) + \lambda_{\text{MI}} \, \text{MI}(r_j, E) \right)}.
\end{equation}
These preference scores are dynamically updated during training and used both for chain selection and for regularisation.
The optimal reasoning chain is selected by maximising the combined signal:
\begin{equation}
r^* = \arg\max_{r_i \in R} \left( \lambda_{\text{IG}} \, \text{IG}(r_i) + \lambda_{\text{MI}} \, \text{MI}(r_i, E) \right),
\end{equation}
where \( \lambda_{\text{IG}} \) and \( \lambda_{\text{MI}} \) are balancing coefficients.

To incorporate the selected reasoning chain into the model, we first encode \( r^* \) into a vector representation. Specifically, we tokenize \( r^* \), pass it through the LLM, and apply mean pooling over its token embeddings to obtain a reasoning embedding \( z_r \in \mathbb{R}^d \). This vector captures the semantic content of the selected rationale in the same latent space as the emotion-aware personality representation \( \tilde{z}_{\text{pers}} \).

We integrate the reasoning vector \( z_r \) with the personality-specific representation \( \tilde{z}_{\text{pers}} \) by applying a lightweight transformation to their concatenation, i.e., \( z_{\text{final}} = \text{MLP}(\text{Concat}(\tilde{z}_{\text{pers}}, z_r)) \), which projects the fused representation back into \( \mathbb{R}^d \). This design allows the model to incorporate high-level inductive signals from the reasoning chain into the personality representation, while preserving the original semantic structure derived from multi-task learning and attention-based interactions.

To encourage confident reasoning selection, we define a reasoning chain entropy loss \( \mathcal{L}_{\text{star}} = - \sum_{i=1}^{n} P(r_i) \log P(r_i) \), which penalizes overly uniform distributions over candidate rationales.

The final training objective combines all loss components, i.e.,
\begin{equation}
\mathcal{L}_{\text{total}} = \lambda_{\text{MTL}} \mathcal{L}_{\text{MTL}} + \lambda_{\text{cross}} \mathcal{L}_{\text{cross}} + \lambda_{\text{star}} \mathcal{L}_{\text{star}},
\end{equation}
where each \( \lambda_{(\cdot)} \) controls the contribution of its corresponding loss term. The resulting fused representation \( z_{\text{final}} \) is then used to compute the final personality prediction \( \hat{y}_{\text{pers}} \).

\section{Experiments and Results}

This section details the benchmark datasets used, experimental settings, and the significant results obtained to determine whether EmoPerso outperforms recent robust models through rigorous evaluation of the effectiveness of EmoPerso on personality detection tasks. We further perform a comprehensive ablation study and visualisation-based analysis to assess the contribution of each component, followed by an in-depth qualitative analysis.

\begin{table}[tbp]
\caption{Statistics on the quantity and class distribution for the Kaggle and Pandora datasets.}
\centering
\scalebox{0.88}
{
    \centering
    \begin{tabular}{l|l|c|c|c}
        \hline
        \textbf{Dataset} & \textbf{Types} & \textbf{Train} & \textbf{Validation} & \textbf{Test} \\
        \hline
        \multirow{5}{*}{\textbf{Kaggle}} & I/E & 1194 / 4011 & 409 / 1326 & 396 / 1339 \\
        & S/N & 610 / 4478 & 222 / 1513 & 248 / 1487 \\
        & T/F & 2410 / 2795 & 791 / 944 & 780 / 955 \\
        & P/J & 2109 / 3096 & 672 / 1063 & 653 / 1082 \\
        & Posts & 246794 & 82642 & 82152 \\
        \hline
        \multirow{5}{*}{\textbf{Pandora}} & I/E & 1162 / 4278 & 386 / 1427 & 377 / 1437 \\
        & S/N & 727 / 4830 & 208 / 1605 & 210 / 1604 \\
        & T/F & 3549 / 1891 & 1120 / 693 & 1182 / 632 \\
        & P/J & 2229 / 3211 & 770 / 1043 & 758 / 1056 \\
        & Posts & 523534 & 173005 & 174080 \\
        \hline
    \end{tabular}
    }
    \label{tab:data-statistics}
    \vspace{-0.075in}
\end{table}

\begin{table*}[t]
\caption{Comparison of EmoPerso with state-of-the-art baselines on the Kaggle and Pandora datasets. Results are reported using Macro-F1 (\%) scores across the four MBTI dimensions and the overall average (Avg).}
    \centering
    \scalebox{0.85}
    {
    \begin{tabular}{l|cccc|c|cccc|c}
        \hline
        \multirow{2}{*}{\textbf{Methods}} & \multicolumn{5}{c|}{\textbf{Kaggle}} & \multicolumn{5}{c}{\textbf{Pandora}} \\
        \cline{2-11}
        & I/E & S/N & T/F & P/J & \textbf{Avg} & I/E & S/N & T/F & P/J & \textbf{Avg} \\
        \hline
        AttRCNN & 59.74 & 64.08 & 78.77 & 66.44 & 67.25 & 48.55 & 56.19 & 64.39 & 57.26 & 56.60 \\
        SN+Attn & 65.43 & 62.15 & 78.05 & 63.92 & 67.39 & 56.98 & 54.78 & 60.95 & 54.81 & 56.88 \\
        Transformer-MD & 66.08 & 69.10 & 79.19 & 67.50 & 70.47 & 55.26 & 58.77 & 69.26 & 60.90 & 61.05 \\
        PQ-Net & 68.94 & 67.65 & 79.12 & 69.57 & 71.32 & 57.07 & 55.26 & 65.64 & 58.74 & 59.18 \\
        TrigNet & 69.54 & 67.17 & 79.06 & 67.69 & 70.86 & 56.69 & 55.57 & 66.38 & 57.27 & 58.98 \\
        PS-GCN & 70.52 & 65.73 & 70.51 & 67.13 & 68.47 & 59.12 & 54.88 & 67.35 & 58.62 & 59.49 \\
        D-DGCN & 69.52 & 67.19 & 80.53 & 68.16 & 71.35 & 59.98 & 55.52 & 70.53 & 59.56 & 61.40 \\
        DEN & 69.95 & 66.39 & 80.65 & 69.02 & 71.50 & 60.86 & 57.74 & 71.64 & 59.17 & 62.35 \\
        MvP & 67.68 & 69.89 & 80.99 & 68.32 & 71.72 & 60.08 & 56.99 & 69.12 & 61.19 & 61.85 \\
        PsyCoT & 66.56 & 61.70 & 74.80 & 57.83 & 65.22 & 60.91 & 57.12 & 66.45 & 53.34 & 59.45 \\
        TAE & 70.90 & 66.21 & 81.17 & 70.20 & 72.07 & 62.57 & 61.01 & 69.28 & 59.34 & 63.05 \\
        \hline
        \textbf{EmoPerso} & \textbf{80.05} & \textbf{79.27} & \textbf{87.03} & \textbf{77.91} & \textbf{81.07} & \textbf{66.84} & \textbf{68.15} & \textbf{71.90} & \textbf{67.51} & \textbf{68.60} \\
        \hline
    \end{tabular}
    }
    \label{tab:macro-f1}
\end{table*}

\subsection{Experimental Setup}

\noindent \textbf{Datasets:} To ensure a fair comparison with previous work, we selected the same two datasets, i.e., Kaggle\footnote{\url{https://www.kaggle.com/datasnaek/mbti-type}} and Pandora\footnote{\url{https://psy.takelab.fer.hr/datasets/all}}. The Kaggle dataset is sourced from the PersonalityCafe forum, an online community focused on discussions about personality types. This dataset contains posts from 8,675 users, with each user contributing approximately 45 to 50 posts. The posts cover a variety of topics, including psychology, personal experiences, and everyday discussions. The dataset is labelled according to users’ self-reported MBTI personality dimensions. Pandora is a larger corpus from the Reddit platform, which includes MBTI labels for 9,084 users, primarily extracted from the flairs (short self-descriptions) in MBTI-related subreddits. The number of posts per user ranges from dozens to hundreds, and due to the diversity of the Reddit community, the content covers a broader range of topics. 
Table \ref{tab:data-statistics} presents some statistics of the two datasets.

\noindent \textbf{Implementation Details:} 
We use a frozen DeepSeek-V3 \cite{liu2024deepseek} as the backbone. To ensure cross-task consistency despite differing token lengths, the input sequence is standardized to 2,048 tokens (median) with a hidden size of 4,096. Training employs Adam with learning rate $1\times10^{-3}$. For augmentation, we generate $k=3$ diverse paraphrases per input using prompt-based top-$p$ sampling ($p=0.9$, temperature 1.0), capped at 512 tokens. Contextual completion masks 10\% of tokens, generating up to 20 per span. KL divergence regularization is weighted by 0.1. For MTL, the loss ratio between personality and emotion tasks is 0.7:0.3, optimized with binary cross-entropy, 
reflecting the primary role of personality prediction and the auxiliary role of emotion modelling. Pseudo-labels for emotion are derived from affective heuristics (e.g., adjectives, intensifiers, and punctuation). Classifier heads are two-layer MLPs with ReLU and dropout 0.2. The cross-attention module uses four heads with residual connections and layer normalization. Reasoning chains are generated by the LLM ($\leq 4$ steps), scored by information gain and mutual information from chain-conditioned vs.\ unconditioned probabilities. Preference scores are computed dynamically with softmax for selection and consistency loss. To avoid leakage, words or phrases directly matching personality labels are removed in preprocessing. Data is split 60/20/20 for train/validation/test, with results averaged over ten runs. Training is conducted on a cluster of NVIDIA H200 GPUs.

\begin{table*}[t]
\caption{Ablation results of different component configurations in EmoPerso. Reporting the Macro-F1 scores (\%) on Kaggle and Pandora datasets for four dimensions and overall average (Avg).}
    \centering
    \scalebox{0.85}
    {
    \begin{tabular}{l|cccc|c|cccc|c}
        \hline
        \multirow{2}{*}{\textbf{Components}} & \multicolumn{5}{c|}{\textbf{Kaggle}} & \multicolumn{5}{c}{\textbf{Pandora}} \\
        \cline{2-11}
        & I/E & S/N & T/F & P/J & \textbf{Avg} & I/E & S/N & T/F & P/J & \textbf{Avg} \\
        \hline
        Vanilla DeepSeek-V3 & 65.16 & 59.75 & 77.71 & 64.43 & 66.76 & 60.17 & 55.93 & 65.58 & 60.35 & 60.51 \\
        w/o Emotions  & 69.31 & 73.79 & 76.01 & 73.00 & 73.03 & 58.85 & 65.69 & 63.15 & 65.16 & 63.21 \\
        w/o Generative Mechanism & 74.43 & 71.22 & 84.30 & 76.11 & 76.52 & 61.61 & 60.90 & 70.16 & 66.41 & 64.77 \\
        w/o Paraphrasing & 77.21 & 74.03 & 85.11 & 76.88 & 78.31 & 64.08 & 65.01 & 70.92 & 66.74 & 66.69 \\
        w/o KL Divergence & 79.41 & 78.12 & 86.41 & 77.03 & 80.24 & 66.02 & 66.97 & 71.33 & 66.45 & 67.69 \\
        w/o MTL & 75.60 & 71.21 & 78.08 & 74.48 & 74.84 & 63.44 & 61.91 & 64.78 & 65.69 & 63.96 \\
        w/o Shared Encoder & 76.92 & 76.83 & 83.55 & 75.01 & 78.58 & 64.03 & 65.81 & 69.01 & 65.16 & 66.50 \\
        w/o CrossAttn Mechanism & 76.47 & 77.05 & 78.94 & 71.95 & 76.10 & 63.96 & 66.83 & 65.14 & 62.10 & 64.51 \\
        Replace CrossAttn with Gated Fusion & 78.32 & 77.93 & 82.36 & 74.88 & 78.87 & 65.15 & 67.32 & 68.03 & 63.97 & 66.12 \\
        w/o Emotion Modulation & 76.83 & 77.14 & 80.91 & 74.48 & 77.84 & 64.12 & 66.58 & 68.21 & 63.88 & 65.70 \\
        w/o Reasoning Chains & 74.70 & 75.03 & 78.82 & 73.15 & 75.42 & 63.04 & 64.78 & 65.51 & 63.27 & 64.15 \\
        Replace STaR with CoT Templates & 76.43 & 76.81 & 81.90 & 75.56 & 77.68 & 64.70 & 65.85 & 68.10 & 64.98 & 65.91 \\
        w/o IG and MI & 78.73 & 77.12 & 85.41 & 75.58 & 79.21 & 65.47 & 66.15 & 70.69 & 66.22 & 67.13 \\
        Replace DeepSeek-V3 with GPT-4o & \textbf{81.12} & 78.23 & 84.45 & \textbf{79.21} & 80.75 & \textbf{68.45} & 67.98 & 70.10 & \textbf{69.77} & \textbf{69.08} \\
        \hline
        \textbf{EmoPerso} & 80.05 & \textbf{79.27} & \textbf{87.03} & 77.91 & \textbf{81.07} & 66.84 & \textbf{68.15} & \textbf{71.90} & 67.51 & 68.60 \\
        \hline
    \end{tabular}
    }
    \label{tab:ablation}
\end{table*}

\noindent \textbf{Evaluation Metrics:} Macro-F1 has been widely used in previous studies and has become the standard evaluation metric for this task. Therefore, we adhere to this convention and use Macro-F1 to ensure consistency with prior work \cite{hu2024llm, yang2023orders, zhu2024integrating}.

\noindent \textbf{Comparative Models:}
We selected diverse sophisticated architectures. AttRCNN \cite{xue2018deep} integrates attention into an RCNN structure with a CNN-Inception module for robust feature extraction. SN+Attn \cite{lynn2020hierarchical} uses a Sequence Network with dual attention at message and word levels to enhance signal relevance. Transformer-MD \cite{yang2021multi} employs Transformer-XL and memory mechanisms for disorder-agnostic post integration with dimension-specific attention. PQ-Net \cite{yang2021learning} fuses psychological questionnaires and user text via cross-attention to capture personality cues. TrigNet \cite{yang2021psycholinguistic} constructs a heterogeneous tripartite graph with flow graph attention (GAT) for psycholinguistic integration. PS-GCN \cite{liu2024ps} merges psycholinguistic knowledge graphs and sentiment semantics via GCN and multi-head attention. D-DGCN \cite{yang2023orders} dynamically builds graph structures, integrating multiple posts disorder-agnostically. DEN \cite{zhu2024enhancing} models long-term personality traits with GCN, short-term states with BERT, and enhances both via bidirectional interaction. MvP \cite{zhu2024integrating} employs a multi-view Mixture-of-Experts with consistency regularization to integrate diverse perspectives of user posts. PsyCoT \cite{yang2023psycot} structures psychological questionnaires as a reasoning chain, using multi-turn dialogue prompting for LLM-based scoring. TAE \cite{hu2024llm} leverages LLM-generated text augmentation and label explanations, applying contrastive learning to improve psycholinguistic representation.

\subsection{Overall Results}
The comparison of Macro-F1 scores between our EmoPerso and the baseline models is presented in Table \ref{tab:macro-f1}. EmoPerso achieved the highest performance across all four dimensions as well as the overall average. On the Kaggle dataset, EmoPerso attained an average score of 81.07\%, outperforming the best existing model, TAE, by 9.00\%. This marks the first time that a personality detection model has surpassed 80\% on this dataset, establishing a new milestone for future research. Similarly, both the I/E and S/N dimensions have reached approximately 80\% for the first time. On the Pandora dataset, EmoPerso achieved an average score of 68.60\%, surpassing TAE by 5.55\%. Notably, compared to the baselines, EmoPerso demonstrated significant improvements under severe class imbalance, particularly reducing the gap between T/F and the other three dimensions.

\subsection{Ablation Study}

The results of the ablation study are shown in Table~\ref{tab:ablation}. First, we evaluate the performance of Vanilla DeepSeek-V3, which does not incorporate any additional optimizations. The results show a significant drop across all four dimensions and the overall average on both the Kaggle and Pandora datasets. This suggests that the designs introduced in EmoPerso are both necessary and effective, systematically addressing the base model's limitations in emotional understanding and interaction-driven reasoning.

\begin{figure}[htbp]
    \centering
    \includegraphics[scale=0.080]{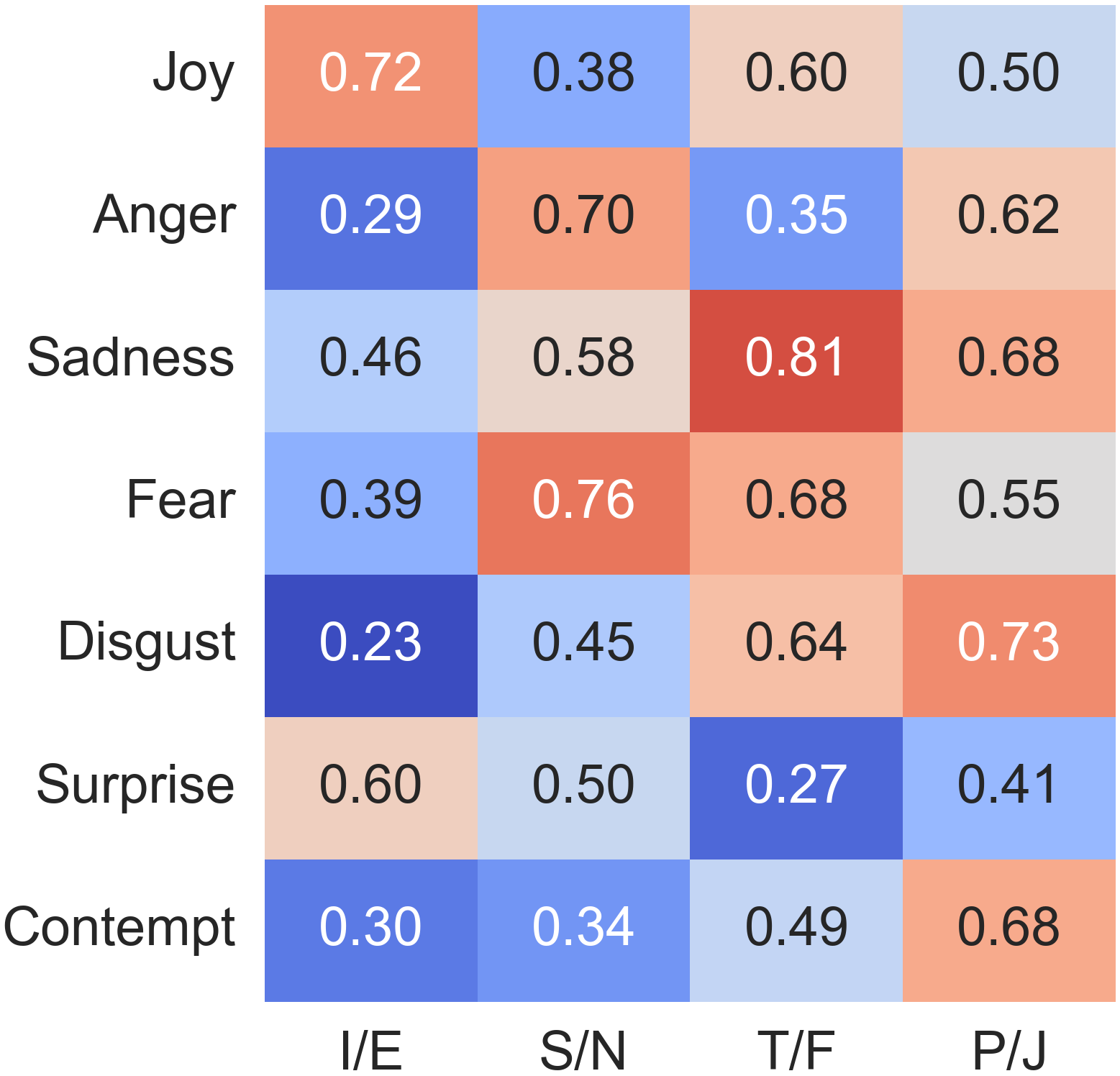}%
    \includegraphics[scale=0.080]{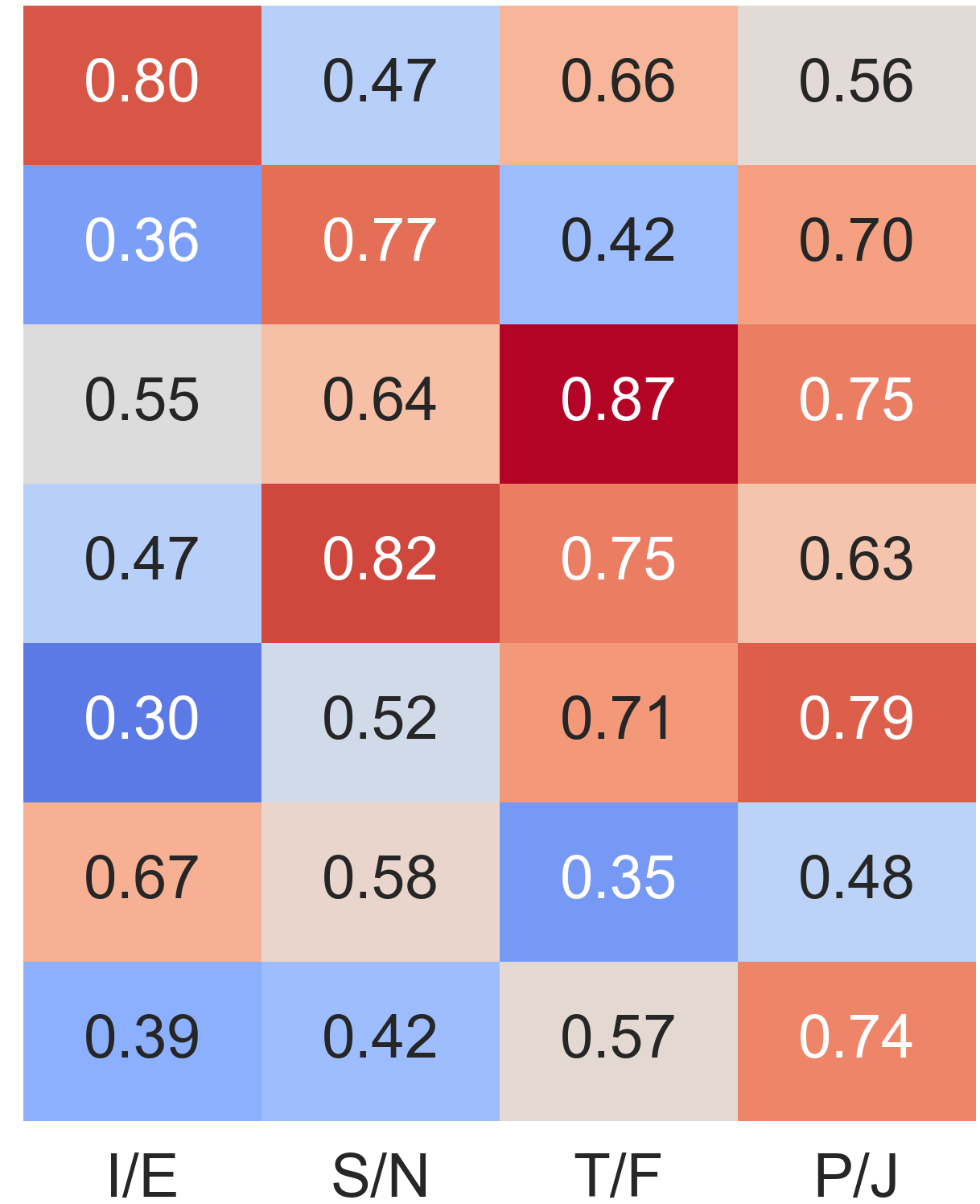}
    \caption{Emotion contribution to prediction on the Kaggle (left) and Pandora (right) datasets.}
    \label{fig:heatmap}
\end{figure}

Next, we remove the emotion features, a key inspiration behind EmoPerso’s design. This results in a substantial drop of 8.04\% on Kaggle and 5.39\% on Pandora, making it the most impactful component in our ablation study. The reason for this is that emotion information serves as a crucial auxiliary signal for EmoPerso, and its absence deprives the model of its most essential enhancement. 
As shown in Figure~\ref{fig:heatmap}, we use a heatmap to quantify the impact of different emotions on MBTI dimensions. The importance score for each emotion is derived from two sources: (1) its influence on the emotion-conditioned attention weights during personality representation refinement, and (2) its contribution to the selection of reasoning chains, measured by the integrated information gain and mutual information scores. The heatmap visualizes how different emotions differentially affect the four MBTI dimensions.

Removing the entire generative mechanism, including style-conditioned paraphrasing and contextual feature completion, leads to a substantial performance drop of 4.55\% on the Kaggle dataset and 3.83\% on the Pandora dataset. This result confirms the importance of personality-related text augmentation in addressing data sparsity and enhancing semantic diversity. 
When the style-conditioned paraphrasing component alone is removed, the performance decreases by 2.76\% on Kaggle and 1.91\% on Pandora. This finding highlights the contribution of paraphrasing to constructing more adaptive and expressive individual communication styles. The impact of removing KL divergence regularization is minimal. In its absence, slight semantic drift \cite{leonardi2024contextual} is observed due to the lack of alignment between original and augmented sequences. 

\begin{figure}[htbp]
    \centering
    \includegraphics[scale=0.09]{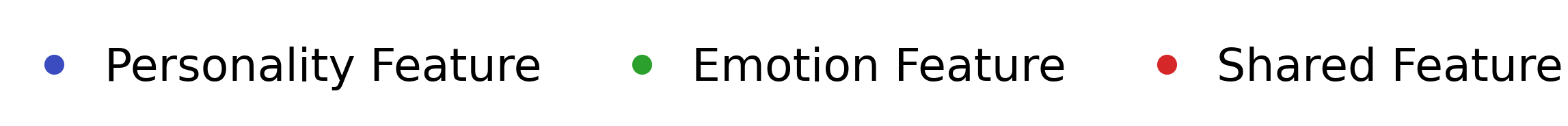}
    \includegraphics[scale=0.08]{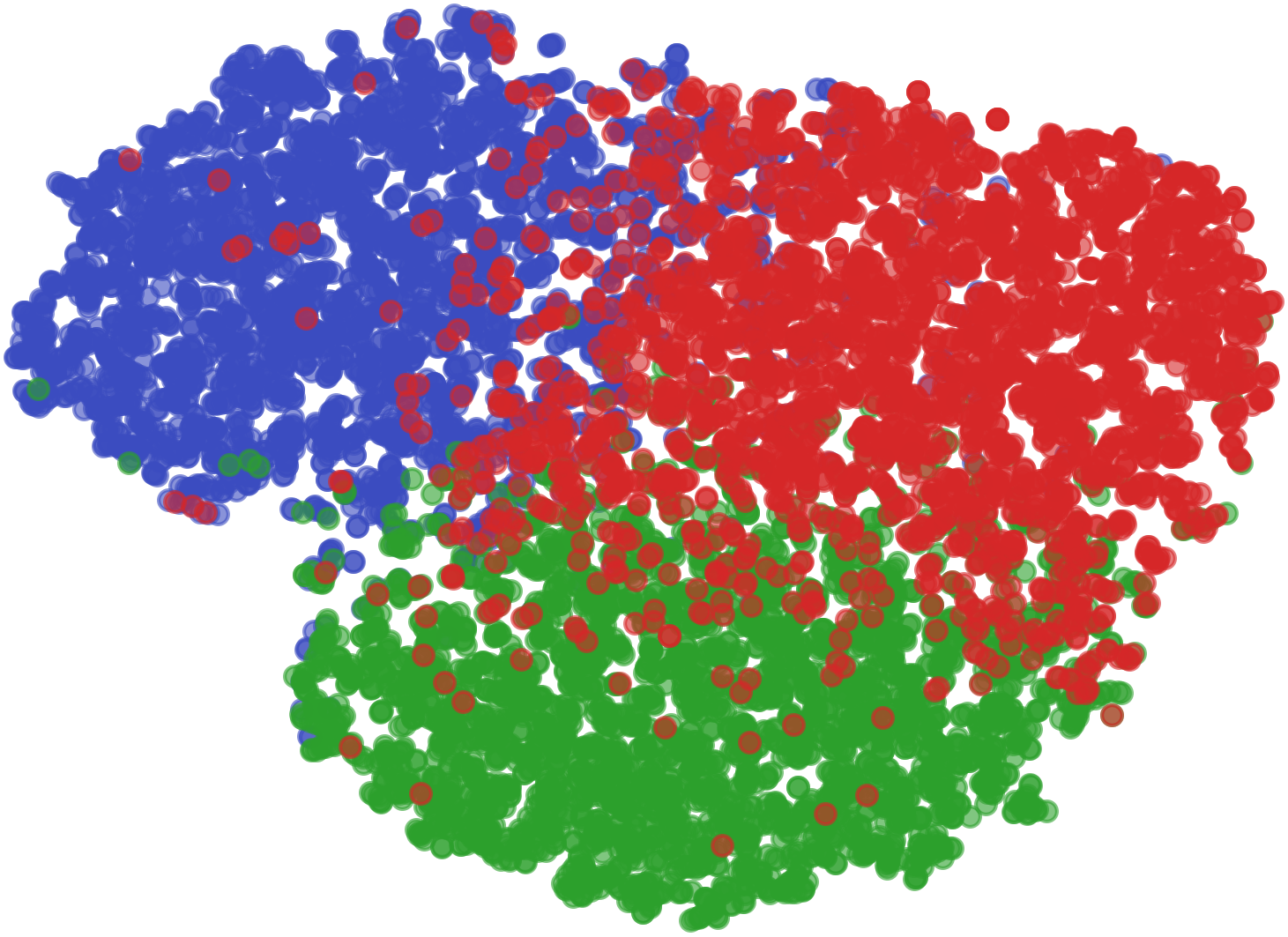}%
    \hspace{0.05cm}
    \includegraphics[scale=0.08]{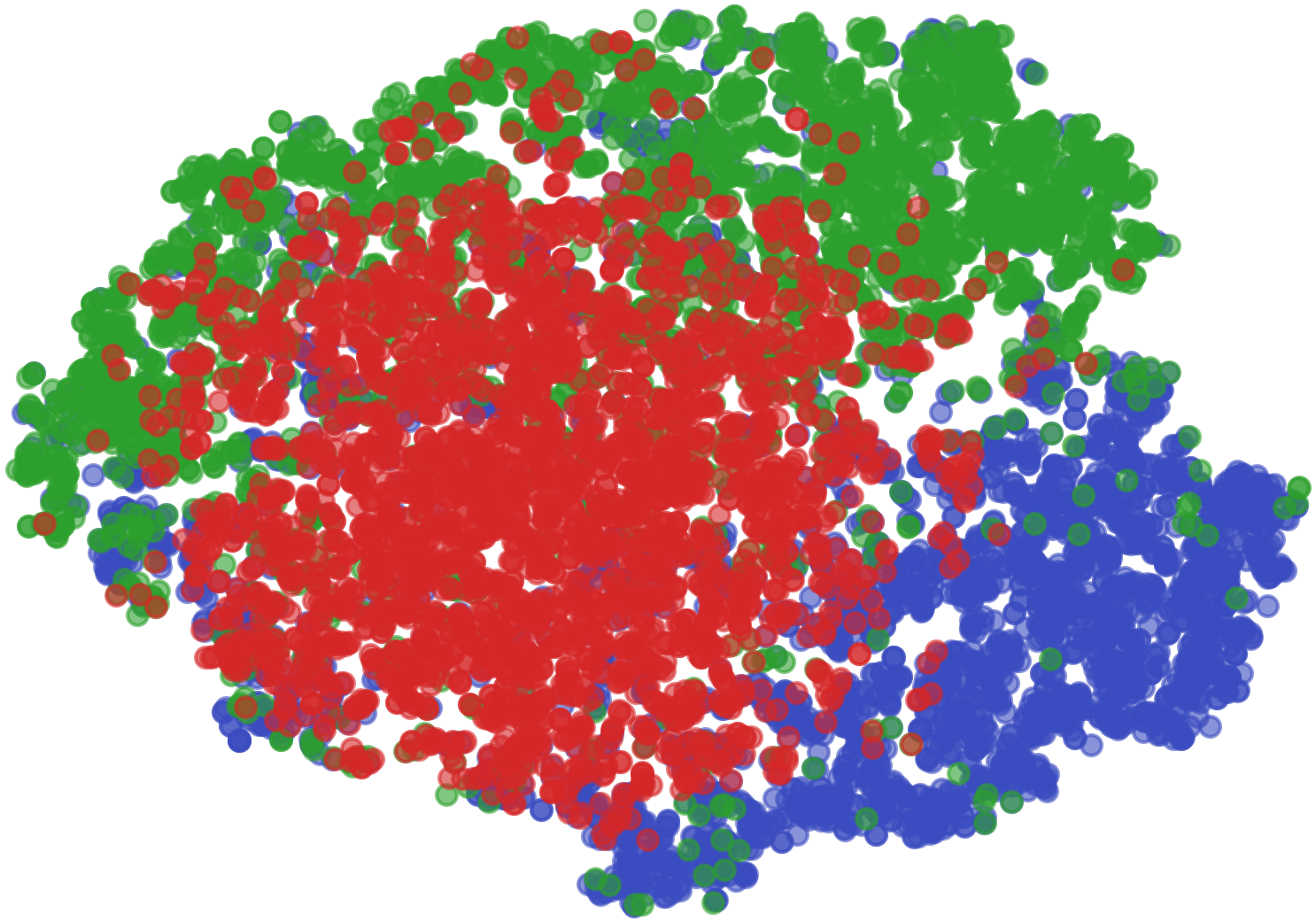}
    \caption{The t-SNE projection visualizes the shared features between personality and emotion under MTL, tested on the Kaggle (left) and Pandora (right).}
    \label{fig:share}
\end{figure}

Following this, we eliminated the MTL designed to optimise emotion--personality interaction, which resulted in the loss of shared representations and the benefits of joint optimisation.
Figure~\ref{fig:share} uses t-SNE on the output representations from the personality head, emotion head, and shared latent vector after multi-task training, revealing substantial overlap between emotion and personality features and thus indicating rich shared representations. The visualization shows that MTL effectively captures and reinforces shared patterns, resulting in more cohesive clustering and better discriminability in the latent space. Additionally, we replaced the shared encoder with two separate encoders for the personality and emotion tasks. This modification introduces a hard separation between the two representation spaces, causing the model to lose the ability to transfer low-level linguistic and emotional cues across tasks.

To evaluate the effect of fine-grained interaction modelling between personality and emotion, we first removed the cross-attention mechanism. This results in a significant performance drop of 4.97\% on Kaggle and 4.09\% on Pandora, confirming that enabling the personality representation to re-attend over the token sequence is critical for integrating emotion-conditioned cues. 
We then replaced the cross-attention mechanism with a gated fusion strategy, where token representations are modulated using a global gating vector derived from both personality and emotion signals. While this alternative may reduces computational complexity, it leads to moderate performance decline. 
The result suggests that although gated fusion partially retains task-level interaction, it lacks the token-level selectivity and subspace diversity provided by multi-head attention. Finally, we removed the emotion modulation component leads to a performance drop of 3.23\% on Kaggle and 2.90\% on Pandora. These findings highlight the role of emotion-guided token weighting in dynamically amplifying psychologically salient cues.

\begin{figure}[htbp]
    \centering
    \includegraphics[scale=0.078]{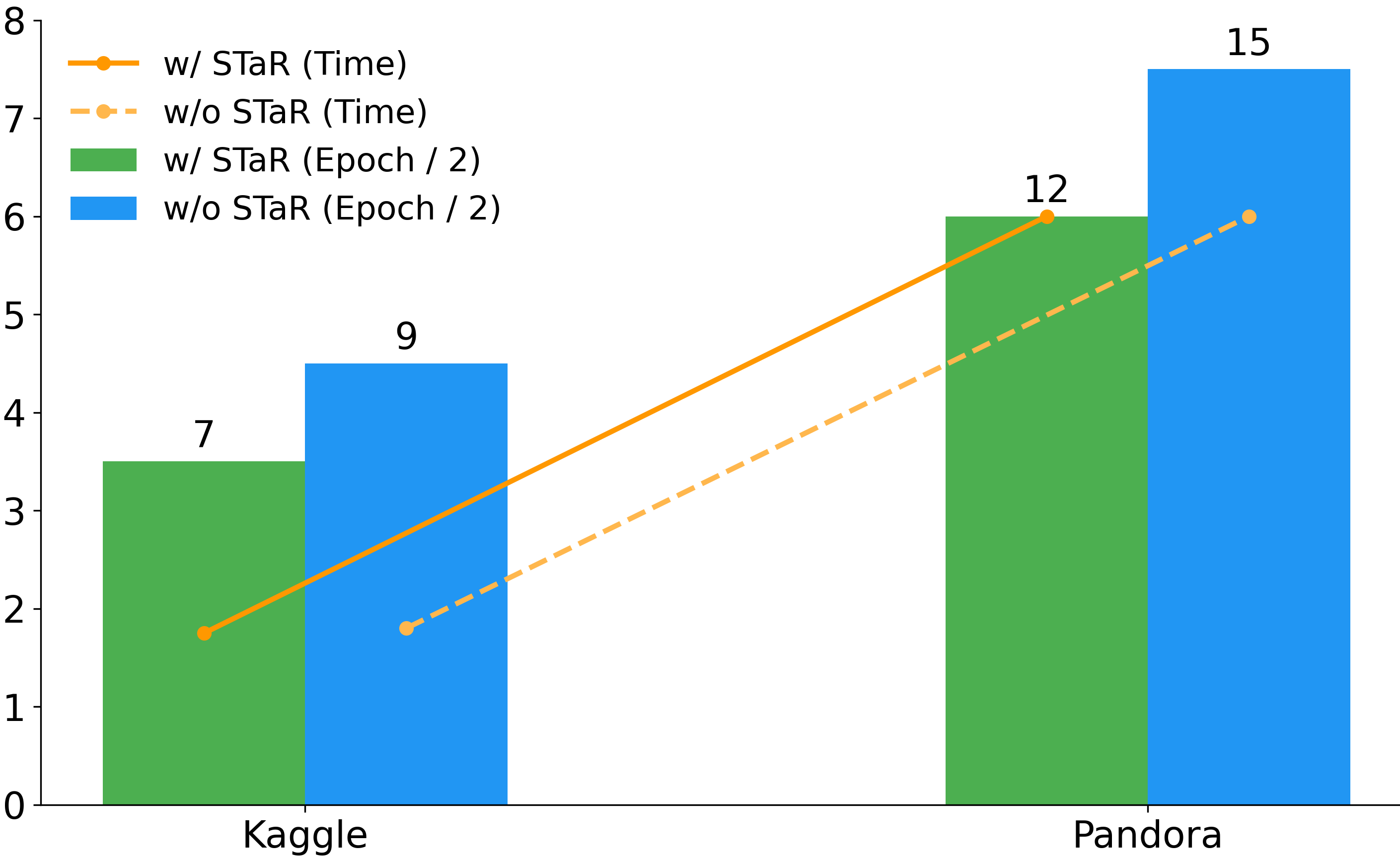} %
    \caption {Comparison of the best training epoch (scaled) and total training time with and without the STaR reasoning module across two datasets. Each bar indicates the optimal number of epochs (halved for visualization), and the lines represent total training time in hours.}
    \label{fig:STaR}
\end{figure}

In the case of reasoning chains, eliminating the STaR module also causes a noticeable performance drop, confirming its critical role in enhancing the model’s reasoning ability. Figure~\ref{fig:STaR} compares the best training epoch (scaled) and total training time (in hours) for EmoPerso with and without STaR on the Kaggle and Pandora datasets. With STaR, the model achieves optimal performance in fewer epochs while maintaining comparable total training time. This suggests that STaR not only enhances personality inference through deeper reasoning but also accelerates the convergence process by guiding the model toward more informative and abstract patterns.
To further probe the quality of reasoning, we replaced STaR with manually crafted Chain-of-Thought (CoT) templates \cite{wei2022chain}, which are fixed prompting patterns designed to elicit step-by-step personality-related reasoning. For example, given a post, a CoT template might produce a generic rationale such as: ``The user expresses frustration, which suggests emotional sensitivity, and therefore may lean toward the Feeling (F) trait.'' This substitution results in moderate performance degradation compared to full STaR, but still performs better than completely removing the reasoning module. The result suggests that while CoT templates can provide basic interpretability, they fail to capture individualized, context-specific reasoning paths. Furthermore, when removing the IG and MI-based reasoning chain selection mechanism, the model still benefits from the existence of reasoning chains, but exhibits a moderate performance drop. This highlights that not all reasoning chains contribute equally to personality inference, and that selecting chains carrying the most informative relational signals is essential for fully leveraging the reasoning process.

Finally, we compare replacing DeepSeek-V3 with GPT-4o\footnote{
Note that although the closed nature of GPT-4o restricts components such as token-level KL divergence and self-supervised optimisation, these are approximated with inference-only strategies or replaced by compatible alternatives, enabling evaluation under a similar inference architecture.} 
as the backbone model and find their performance comparable. On Kaggle, DeepSeek-V3-based EmoPerso slightly outperforms GPT-4o-based EmoPerso on average, whereas on Pandora, the GPT-4o variant surpasses DeepSeek-V3 version. 
Interestingly, DeepSeek-V3 performs better on S/N and T/F, while GPT-4o excels on I/E and P/J, possibly because DeepSeek-V3 captures structured reasoning patterns, whereas GPT-4o better models conversational and social traits, directly influencing these dimensions.

\section{Conclusion}
This paper proposes EmoPerso, a novel self-supervised framework for joint emotion-personality modelling leveraging LLMs. By integrating generative mechanisms, MTL, and cross-attention, the framework facilitates deep interactions between personality and emotion, while STaR enhances the model’s emotion-conditioned reasoning capabilities. Experiments on Kaggle and Pandora show EmoPerso surpasses state-of-the-art models, and ablation studies confirm the importance of each core component. EmoPerso also holds potential for real-world applications, such as mental health screening, bias-aware assessment, personalized marketing, and AI interaction systems. Future work includes multilingual adaptation, advanced generative reasoning, and modality extension.

\section{Limitations and Ethical Considerations}
While EmoPerso demonstrated promising outcomes, our experiments revealed several constraints. The current datasets suffer from label imbalance due to the scarcity of certain personality dimensions and biases in the data collection process. To enhance model's generalization ability, more diverse and balanced real-world datasets are required, beyond the use of synthetically generated data alone. Moreover, the prevalence of social media bots further diminishes the reliability of the data. 

Automated personality recognition entails a range of ethical challenges, including cultural bias, discrimination, and breaches of privacy. Strict adherence to ethical guidelines is essential to ensure fairness and safeguard data confidentiality. Most existing models are predominantly trained on English data, which significantly limits their applicability across different languages and cultural contexts. In addition, personality detection and classification based on emotions and spoken language are inherently complex and prone to biases against specific cultural or ethnic groups. The reliance on potentially biased data sources, such as self-reports and social media content, may undermine the credibility and practical value of applying the model in real-world scenarios.

\begin{acks}
This work was supported by the Alan Turing Institute and Singapore's DSO National Laboratories under a grant on improving multimodal misinformation detection through affective analysis.
\end{acks}

\clearpage

\section*{GenAI Usage Disclosure}
This work employs DeepSeek-V3 as the backbone to perform controlled data augmentation (see Section~\ref{sec:dataaug}), such as paraphrase generation and contextual feature completion, as part of our model’s self-supervised training pipeline. These generative mechanisms were strictly used to enrich the training data and were evaluated for semantic consistency and task relevance. All model design, implementation, analysis, and writing decisions were made by the authors. No generative AI tools were listed as co-authors or assumed intellectual responsibility for any part of this work.

\bibliographystyle{ACM-Reference-Format}
\balance
\bibliography{main}


\begin{thebibliography}{59}


\ifx \showCODEN    \undefined \def \showCODEN     #1{\unskip}     \fi
\ifx \showISBNx    \undefined \def \showISBNx     #1{\unskip}     \fi
\ifx \showISBNxiii \undefined \def \showISBNxiii  #1{\unskip}     \fi
\ifx \showISSN     \undefined \def \showISSN      #1{\unskip}     \fi
\ifx \showLCCN     \undefined \def \showLCCN      #1{\unskip}     \fi
\ifx \shownote     \undefined \def \shownote      #1{#1}          \fi
\ifx \showarticletitle \undefined \def \showarticletitle #1{#1}   \fi
\ifx \showURL      \undefined \def \showURL       {\relax}        \fi
\providecommand\bibfield[2]{#2}
\providecommand\bibinfo[2]{#2}
\providecommand\natexlab[1]{#1}
\providecommand\showeprint[2][]{arXiv:#2}

\bibitem[Achiam et~al\mbox{.}(2023)]%
        {achiam2023gpt}
\bibfield{author}{\bibinfo{person}{Josh Achiam}, \bibinfo{person}{Steven
  Adler}, \bibinfo{person}{Sandhini Agarwal}, \bibinfo{person}{Lama Ahmad},
  \bibinfo{person}{Ilge Akkaya}, \bibinfo{person}{Florencia~Leoni Aleman},
  \bibinfo{person}{Diogo Almeida}, \bibinfo{person}{Janko Altenschmidt},
  \bibinfo{person}{Sam Altman}, \bibinfo{person}{Shyamal Anadkat},
  {et~al\mbox{.}}} \bibinfo{year}{2023}\natexlab{}.
\newblock \showarticletitle{Gpt-4 technical report}.
\newblock \bibinfo{journal}{\emph{arXiv preprint arXiv:2303.08774}}
  (\bibinfo{year}{2023}).
\newblock


\bibitem[Aru et~al\mbox{.}(2023)]%
        {aru2023mind}
\bibfield{author}{\bibinfo{person}{Jaan Aru}, \bibinfo{person}{Aqeel Labash},
  \bibinfo{person}{Oriol Corcoll}, {and} \bibinfo{person}{Raul Vicente}.}
  \bibinfo{year}{2023}\natexlab{}.
\newblock \showarticletitle{Mind the gap: Challenges of deep learning
  approaches to theory of mind}.
\newblock \bibinfo{journal}{\emph{Artificial Intelligence Review}}
  \bibinfo{volume}{56}, \bibinfo{number}{9} (\bibinfo{year}{2023}),
  \bibinfo{pages}{9141--9156}.
\newblock


\bibitem[Buba{\v{s}}(2024)]%
        {bubavs2024use}
\bibfield{author}{\bibinfo{person}{Goran Buba{\v{s}}}.}
  \bibinfo{year}{2024}\natexlab{}.
\newblock \showarticletitle{The use of GPT-4o and Other Large Language Models
  for the Improvement and Design of Self-Assessment Scales for Measurement of
  Interpersonal Communication Skills}.
\newblock \bibinfo{journal}{\emph{arXiv preprint arXiv:2409.14050}}
  (\bibinfo{year}{2024}).
\newblock


\bibitem[Butt et~al\mbox{.}(2025)]%
        {butt2025interpretation}
\bibfield{author}{\bibinfo{person}{Sabur Butt}, \bibinfo{person}{Grigori
  Sidorov}, {and} \bibinfo{person}{Alexander Gelbukh}.}
  \bibinfo{year}{2025}\natexlab{}.
\newblock \showarticletitle{Interpretation of Myers--Briggs Type Indicator
  personality profiles based on ambivert continuum scale}.
\newblock \bibinfo{journal}{\emph{Expert Systems with Applications}}
  \bibinfo{volume}{264} (\bibinfo{year}{2025}), \bibinfo{pages}{125689}.
\newblock


\bibitem[Cascio~Rizzo et~al\mbox{.}(2023)]%
        {cascio2023sensory}
\bibfield{author}{\bibinfo{person}{Giovanni~Luca Cascio~Rizzo},
  \bibinfo{person}{Jonah Berger}, \bibinfo{person}{Matteo De~Angelis}, {and}
  \bibinfo{person}{Rumen Pozharliev}.} \bibinfo{year}{2023}\natexlab{}.
\newblock \showarticletitle{How sensory language shapes influencer’s impact}.
\newblock \bibinfo{journal}{\emph{Journal of Consumer Research}}
  \bibinfo{volume}{50}, \bibinfo{number}{4} (\bibinfo{year}{2023}),
  \bibinfo{pages}{810--825}.
\newblock


\bibitem[Cheng and Shi(2025)]%
        {cheng2025hierarchical}
\bibfield{author}{\bibinfo{person}{Quan Cheng} {and} \bibinfo{person}{Wenwan
  Shi}.} \bibinfo{year}{2025}\natexlab{}.
\newblock \showarticletitle{Hierarchical multi-label text classification of
  tourism resources using a label-aware dual graph attention network}.
\newblock \bibinfo{journal}{\emph{Information Processing \& Management}}
  \bibinfo{volume}{62}, \bibinfo{number}{1} (\bibinfo{year}{2025}),
  \bibinfo{pages}{103952}.
\newblock


\bibitem[Cogan et~al\mbox{.}(2023)]%
        {cogan2023deterministic}
\bibfield{author}{\bibinfo{person}{Dan Cogan}, \bibinfo{person}{Zu-En Su},
  \bibinfo{person}{Oded Kenneth}, {and} \bibinfo{person}{David Gershoni}.}
  \bibinfo{year}{2023}\natexlab{}.
\newblock \showarticletitle{Deterministic generation of indistinguishable
  photons in a cluster state}.
\newblock \bibinfo{journal}{\emph{Nature Photonics}} \bibinfo{volume}{17},
  \bibinfo{number}{4} (\bibinfo{year}{2023}), \bibinfo{pages}{324--329}.
\newblock


\bibitem[Cui et~al\mbox{.}(2025)]%
        {cui2025generalized}
\bibfield{author}{\bibinfo{person}{Jiequan Cui}, \bibinfo{person}{Beier Zhu},
  \bibinfo{person}{Qingshan Xu}, \bibinfo{person}{Zhuotao Tian},
  \bibinfo{person}{Xiaojuan Qi}, \bibinfo{person}{Bei Yu},
  \bibinfo{person}{Hanwang Zhang}, {and} \bibinfo{person}{Richang Hong}.}
  \bibinfo{year}{2025}\natexlab{}.
\newblock \showarticletitle{Generalized Kullback-Leibler Divergence Loss}.
\newblock \bibinfo{journal}{\emph{arXiv preprint arXiv:2503.08038}}
  (\bibinfo{year}{2025}).
\newblock


\bibitem[Ekman(1992)]%
        {Ekman1992}
\bibfield{author}{\bibinfo{person}{P. Ekman}.} \bibinfo{year}{1992}\natexlab{}.
\newblock \showarticletitle{Are there basic emotions?}
\newblock \bibinfo{journal}{\emph{Psychological Review}} \bibinfo{volume}{99},
  \bibinfo{number}{3} (\bibinfo{year}{1992}), \bibinfo{pages}{550--553}.
\newblock


\bibitem[Grattafiori et~al\mbox{.}(2024)]%
        {grattafiori2024llama}
\bibfield{author}{\bibinfo{person}{Aaron Grattafiori},
  \bibinfo{person}{Abhimanyu Dubey}, \bibinfo{person}{Abhinav Jauhri},
  \bibinfo{person}{Abhinav Pandey}, \bibinfo{person}{Abhishek Kadian},
  \bibinfo{person}{Ahmad Al-Dahle}, \bibinfo{person}{Aiesha Letman},
  \bibinfo{person}{Akhil Mathur}, \bibinfo{person}{Alan Schelten},
  \bibinfo{person}{Alex Vaughan}, {et~al\mbox{.}}}
  \bibinfo{year}{2024}\natexlab{}.
\newblock \showarticletitle{The llama 3 herd of models}.
\newblock \bibinfo{journal}{\emph{arXiv preprint arXiv:2407.21783}}
  (\bibinfo{year}{2024}).
\newblock


\bibitem[Guerra et~al\mbox{.}(2022)]%
        {guerra2022datasets}
\bibfield{author}{\bibinfo{person}{Jorge~Luis Guerra}, \bibinfo{person}{Carlos
  Catania}, {and} \bibinfo{person}{Eduardo Veas}.}
  \bibinfo{year}{2022}\natexlab{}.
\newblock \showarticletitle{Datasets are not enough: Challenges in labeling
  network traffic}.
\newblock \bibinfo{journal}{\emph{Computers \& Security}}
  \bibinfo{volume}{120} (\bibinfo{year}{2022}), \bibinfo{pages}{102810}.
\newblock


\bibitem[Hu et~al\mbox{.}(2024)]%
        {hu2024llm}
\bibfield{author}{\bibinfo{person}{Linmei Hu}, \bibinfo{person}{Hongyu He},
  \bibinfo{person}{Duokang Wang}, \bibinfo{person}{Ziwang Zhao},
  \bibinfo{person}{Yingxia Shao}, {and} \bibinfo{person}{Liqiang Nie}.}
  \bibinfo{year}{2024}\natexlab{}.
\newblock \showarticletitle{LLM vs Small Model? Large Language Model Based Text
  Augmentation Enhanced Personality Detection Model}. In
  \bibinfo{booktitle}{\emph{Proceedings of the AAAI Conference on Artificial
  Intelligence}}, Vol.~\bibinfo{volume}{38}. \bibinfo{pages}{18234--18242}.
\newblock


\bibitem[Jaradat et~al\mbox{.}(2024)]%
        {jaradat2024multitask}
\bibfield{author}{\bibinfo{person}{Shadi Jaradat}, \bibinfo{person}{Richi
  Nayak}, \bibinfo{person}{Alexander Paz}, \bibinfo{person}{Huthaifa~I Ashqar},
  {and} \bibinfo{person}{Mohammad Elhenawy}.} \bibinfo{year}{2024}\natexlab{}.
\newblock \showarticletitle{Multitask learning for crash analysis: A fine-tuned
  llm framework using twitter data}.
\newblock \bibinfo{journal}{\emph{Smart Cities}} \bibinfo{volume}{7},
  \bibinfo{number}{5} (\bibinfo{year}{2024}), \bibinfo{pages}{2422--2465}.
\newblock


\bibitem[Jim et~al\mbox{.}(2024)]%
        {jim2024recent}
\bibfield{author}{\bibinfo{person}{Jamin~Rahman Jim},
  \bibinfo{person}{Md~Apon~Riaz Talukder}, \bibinfo{person}{Partha Malakar},
  \bibinfo{person}{Md~Mohsin Kabir}, \bibinfo{person}{Kamruddin Nur}, {and}
  \bibinfo{person}{Mohammed~Firoz Mridha}.} \bibinfo{year}{2024}\natexlab{}.
\newblock \showarticletitle{Recent advancements and challenges of NLP-based
  sentiment analysis: A state-of-the-art review}.
\newblock \bibinfo{journal}{\emph{Natural Language Processing Journal}}
  (\bibinfo{year}{2024}), \bibinfo{pages}{100059}.
\newblock


\bibitem[Leonardi et~al\mbox{.}(2024)]%
        {leonardi2024contextual}
\bibfield{author}{\bibinfo{person}{Francois Leonardi}, \bibinfo{person}{Patrick
  Feldman}, \bibinfo{person}{Matthew Almeida}, \bibinfo{person}{William
  Moretti}, {and} \bibinfo{person}{Charles Iverson}.}
  \bibinfo{year}{2024}\natexlab{}.
\newblock \showarticletitle{Contextual feature drift in large language models:
  An examination of adaptive retention across sequential inputs}.
\newblock  (\bibinfo{year}{2024}).
\newblock


\bibitem[Li et~al\mbox{.}(2023)]%
        {li2023uniformer}
\bibfield{author}{\bibinfo{person}{Kunchang Li}, \bibinfo{person}{Yali Wang},
  \bibinfo{person}{Junhao Zhang}, \bibinfo{person}{Peng Gao},
  \bibinfo{person}{Guanglu Song}, \bibinfo{person}{Yu Liu},
  \bibinfo{person}{Hongsheng Li}, {and} \bibinfo{person}{Yu Qiao}.}
  \bibinfo{year}{2023}\natexlab{}.
\newblock \showarticletitle{Uniformer: Unifying convolution and self-attention
  for visual recognition}.
\newblock \bibinfo{journal}{\emph{IEEE Transactions on Pattern Analysis and
  Machine Intelligence}} \bibinfo{volume}{45}, \bibinfo{number}{10}
  (\bibinfo{year}{2023}), \bibinfo{pages}{12581--12600}.
\newblock


\bibitem[Li et~al\mbox{.}(2024a)]%
        {li2024explanation}
\bibfield{author}{\bibinfo{person}{Yuming Li}, \bibinfo{person}{Johnny Chan},
  \bibinfo{person}{Gabrielle Peko}, {and} \bibinfo{person}{David Sundaram}.}
  \bibinfo{year}{2024}\natexlab{a}.
\newblock \showarticletitle{An explanation framework and method for AI-based
  text emotion analysis and visualisation}.
\newblock \bibinfo{journal}{\emph{Decision Support Systems}}
  \bibinfo{volume}{178} (\bibinfo{year}{2024}), \bibinfo{pages}{114121}.
\newblock


\bibitem[Li et~al\mbox{.}(2022)]%
        {li2022multitask}
\bibfield{author}{\bibinfo{person}{Yang Li}, \bibinfo{person}{Amirmohammad
  Kazemeini}, \bibinfo{person}{Yash Mehta}, {and} \bibinfo{person}{Erik
  Cambria}.} \bibinfo{year}{2022}\natexlab{}.
\newblock \showarticletitle{Multitask learning for emotion and personality
  traits detection}.
\newblock \bibinfo{journal}{\emph{Neurocomputing}}  \bibinfo{volume}{493}
  (\bibinfo{year}{2022}), \bibinfo{pages}{340--350}.
\newblock


\bibitem[Li et~al\mbox{.}(2024b)]%
        {li2024eerpd}
\bibfield{author}{\bibinfo{person}{Zheng Li}, \bibinfo{person}{Dawei Zhu},
  \bibinfo{person}{Qilong Ma}, \bibinfo{person}{Weimin Xiong}, {and}
  \bibinfo{person}{Sujian Li}.} \bibinfo{year}{2024}\natexlab{b}.
\newblock \showarticletitle{EERPD: Leveraging Emotion and Emotion Regulation
  for Improving Personality Detection}.
\newblock \bibinfo{journal}{\emph{arXiv preprint arXiv:2406.16079}}
  (\bibinfo{year}{2024}).
\newblock


\bibitem[Liu et~al\mbox{.}(2024a)]%
        {liu2024deepseek}
\bibfield{author}{\bibinfo{person}{Aixin Liu}, \bibinfo{person}{Bei Feng},
  \bibinfo{person}{Bing Xue}, \bibinfo{person}{Bingxuan Wang},
  \bibinfo{person}{Bochao Wu}, \bibinfo{person}{Chengda Lu},
  \bibinfo{person}{Chenggang Zhao}, \bibinfo{person}{Chengqi Deng},
  \bibinfo{person}{Chenyu Zhang}, \bibinfo{person}{Chong Ruan},
  {et~al\mbox{.}}} \bibinfo{year}{2024}\natexlab{a}.
\newblock \showarticletitle{Deepseek-v3 technical report}.
\newblock \bibinfo{journal}{\emph{arXiv preprint arXiv:2412.19437}}
  (\bibinfo{year}{2024}).
\newblock


\bibitem[Liu et~al\mbox{.}(2025)]%
        {liu2025knowledge}
\bibfield{author}{\bibinfo{person}{Wenjie Liu}, \bibinfo{person}{Zhijie Ren},
  {and} \bibinfo{person}{Liang Chen}.} \bibinfo{year}{2025}\natexlab{}.
\newblock \showarticletitle{Knowledge reasoning based on graph neural networks
  with multi-layer top-p message passing and sparse negative sampling}.
\newblock \bibinfo{journal}{\emph{Knowledge-Based Systems}}
  (\bibinfo{year}{2025}), \bibinfo{pages}{113063}.
\newblock


\bibitem[Liu et~al\mbox{.}(2024b)]%
        {liu2024ps}
\bibfield{author}{\bibinfo{person}{Wenjuan Liu}, \bibinfo{person}{Zhengyan
  Sun}, \bibinfo{person}{Subo Wei}, \bibinfo{person}{Shunxiang Zhang},
  \bibinfo{person}{Guangli Zhu}, {and} \bibinfo{person}{Lei Chen}.}
  \bibinfo{year}{2024}\natexlab{b}.
\newblock \showarticletitle{PS-GCN: Psycholinguistic graph and sentiment
  semantic fused graph convolutional networks for personality detection}.
\newblock \bibinfo{journal}{\emph{Connection Science}} \bibinfo{volume}{36},
  \bibinfo{number}{1} (\bibinfo{year}{2024}), \bibinfo{pages}{2295820}.
\newblock


\bibitem[Luo et~al\mbox{.}(2023)]%
        {luo2023self}
\bibfield{author}{\bibinfo{person}{Qing Luo}, \bibinfo{person}{Wei Zeng},
  \bibinfo{person}{Manni Chen}, \bibinfo{person}{Gang Peng},
  \bibinfo{person}{Xiaofeng Yuan}, {and} \bibinfo{person}{Qiang Yin}.}
  \bibinfo{year}{2023}\natexlab{}.
\newblock \showarticletitle{Self-Attention and Transformers: Driving the
  Evolution of Large Language Models}. In \bibinfo{booktitle}{\emph{2023 IEEE
  6th International Conference on Electronic Information and Communication
  Technology (ICEICT)}}. IEEE, \bibinfo{pages}{401--405}.
\newblock


\bibitem[Lynn et~al\mbox{.}(2020)]%
        {lynn2020hierarchical}
\bibfield{author}{\bibinfo{person}{Veronica Lynn}, \bibinfo{person}{Niranjan
  Balasubramanian}, {and} \bibinfo{person}{H~Andrew Schwartz}.}
  \bibinfo{year}{2020}\natexlab{}.
\newblock \showarticletitle{Hierarchical modeling for user personality
  prediction: The role of message-level attention}. In
  \bibinfo{booktitle}{\emph{Proceedings of the 58th annual meeting of the
  association for computational linguistics}}. \bibinfo{pages}{5306--5316}.
\newblock


\bibitem[Mischel and Shoda(1995)]%
        {mischel1995cognitive}
\bibfield{author}{\bibinfo{person}{Walter Mischel} {and}
  \bibinfo{person}{Yuichi Shoda}.} \bibinfo{year}{1995}\natexlab{}.
\newblock \showarticletitle{A cognitive-affective system theory of personality:
  reconceptualizing situations, dispositions, dynamics, and invariance in
  personality structure.}
\newblock \bibinfo{journal}{\emph{Psychological review}} \bibinfo{volume}{102},
  \bibinfo{number}{2} (\bibinfo{year}{1995}), \bibinfo{pages}{246}.
\newblock


\bibitem[Murphy(2024)]%
        {murphy2024artificial}
\bibfield{author}{\bibinfo{person}{Max Murphy}.}
  \bibinfo{year}{2024}\natexlab{}.
\newblock \showarticletitle{Artificial Intelligence and Personality: Large
  Language Models’ Ability to Predict Personality Type}.
\newblock \bibinfo{journal}{\emph{Emerging Media}} (\bibinfo{year}{2024}),
  \bibinfo{pages}{27523543241257291}.
\newblock


\bibitem[Nelson et~al\mbox{.}(2025)]%
        {nelson2025evaluating}
\bibfield{author}{\bibinfo{person}{Benjamin Nelson}, \bibinfo{person}{Ari
  Winbush}, \bibinfo{person}{Steven Siddals}, \bibinfo{person}{John Torous},
  \bibinfo{person}{Nick Allen}, {and} \bibinfo{person}{Matthew Flathers}.}
  \bibinfo{year}{2025}\natexlab{}.
\newblock \showarticletitle{Evaluating the Performance of Large Language Models
  in Identifying Human Facial Emotions: GPT 4o, Gemini 2.0 Experimental, and
  Claude 3.5 Sonnet}.
\newblock  (\bibinfo{year}{2025}).
\newblock


\bibitem[Pliskin et~al\mbox{.}(2020)]%
        {pliskin2020proposing}
\bibfield{author}{\bibinfo{person}{Ruthie Pliskin}, \bibinfo{person}{Anat
  Ruhrman}, {and} \bibinfo{person}{Eran Halperin}.}
  \bibinfo{year}{2020}\natexlab{}.
\newblock \showarticletitle{Proposing a multi-dimensional, context-sensitive
  approach to the study of ideological (a) symmetry in emotion}.
\newblock \bibinfo{journal}{\emph{Current Opinion in Behavioral Sciences}}
  \bibinfo{volume}{34} (\bibinfo{year}{2020}), \bibinfo{pages}{75--80}.
\newblock


\bibitem[Prasanthi and Anuradha(2021)]%
        {prasanthi2021survey}
\bibfield{author}{\bibinfo{person}{J Prasanthi} {and} \bibinfo{person}{G
  Anuradha}.} \bibinfo{year}{2021}\natexlab{}.
\newblock \showarticletitle{SURVEY ON PERSONALITY DETECTION USING DEEP LEARNING
  TECHNIQUES}. In \bibinfo{booktitle}{\emph{2021 6th International Conference
  on Communication and Electronics Systems (ICCES)}}. IEEE,
  \bibinfo{pages}{1--8}.
\newblock


\bibitem[Safdari et~al\mbox{.}(2023)]%
        {safdari2023personality}
\bibfield{author}{\bibinfo{person}{Mustafa Safdari}, \bibinfo{person}{Greg
  Serapio-Garc{\'\i}a}, \bibinfo{person}{Cl{\'e}ment Crepy},
  \bibinfo{person}{Stephen Fitz}, \bibinfo{person}{Peter Romero},
  \bibinfo{person}{Luning Sun}, \bibinfo{person}{Marwa Abdulhai},
  \bibinfo{person}{Aleksandra Faust}, {and} \bibinfo{person}{Maja
  Matari{\'c}}.} \bibinfo{year}{2023}\natexlab{}.
\newblock \showarticletitle{Personality traits in large language models}.
\newblock \bibinfo{journal}{\emph{arXiv preprint arXiv:2307.00184}}
  (\bibinfo{year}{2023}).
\newblock


\bibitem[Shanmukha et~al\mbox{.}(2024)]%
        {shanmukha2024advancing}
\bibfield{author}{\bibinfo{person}{Aditya~G Shanmukha}, \bibinfo{person}{RS
  Shamyuktha}, \bibinfo{person}{S Karan}, \bibinfo{person}{Deepa Gupta}, {and}
  \bibinfo{person}{Suja Palaniswamy}.} \bibinfo{year}{2024}\natexlab{}.
\newblock \showarticletitle{Advancing Personality Detection through Word
  Embedments and Deep Learning: An Examination Using the MBTI Dataset}. In
  \bibinfo{booktitle}{\emph{2024 IEEE Recent Advances in Intelligent
  Computational Systems (RAICS)}}. IEEE, \bibinfo{pages}{1--6}.
\newblock


\bibitem[Shen et~al\mbox{.}(2025a)]%
        {shen2025less}
\bibfield{author}{\bibinfo{person}{Lingzhi Shen}, \bibinfo{person}{Yunfei
  Long}, \bibinfo{person}{Xiaohao Cai}, \bibinfo{person}{Guanming Chen},
  \bibinfo{person}{Imran Razzak}, {and} \bibinfo{person}{Shoaib Jameel}.}
  \bibinfo{year}{2025}\natexlab{a}.
\newblock \showarticletitle{Less but Better: Parameter-Efficient Fine-Tuning of
  Large Language Models for Personality Detection}.
\newblock \bibinfo{journal}{\emph{arXiv preprint arXiv:2504.05411}}
  (\bibinfo{year}{2025}).
\newblock


\bibitem[Shen et~al\mbox{.}(2025b)]%
        {shen2025ll4g}
\bibfield{author}{\bibinfo{person}{Lingzhi Shen}, \bibinfo{person}{Yunfei
  Long}, \bibinfo{person}{Xiaohao Cai}, \bibinfo{person}{Guanming Chen},
  \bibinfo{person}{Yuhan Wang}, \bibinfo{person}{Imran Razzak}, {and}
  \bibinfo{person}{Shoaib Jameel}.} \bibinfo{year}{2025}\natexlab{b}.
\newblock \showarticletitle{Ll4g: Self-supervised dynamic optimization for
  graph-based personality detection}.
\newblock \bibinfo{journal}{\emph{arXiv preprint arXiv:2504.02146}}
  (\bibinfo{year}{2025}).
\newblock


\bibitem[Shen et~al\mbox{.}(2025c)]%
        {shen2025gamed}
\bibfield{author}{\bibinfo{person}{Lingzhi Shen}, \bibinfo{person}{Yunfei
  Long}, \bibinfo{person}{Xiaohao Cai}, \bibinfo{person}{Imran Razzak},
  \bibinfo{person}{Guanming Chen}, \bibinfo{person}{Kang Liu}, {and}
  \bibinfo{person}{Shoaib Jameel}.} \bibinfo{year}{2025}\natexlab{c}.
\newblock \showarticletitle{Gamed: Knowledge adaptive multi-experts decoupling
  for multimodal fake news detection}. In \bibinfo{booktitle}{\emph{Proceedings
  of the Eighteenth ACM International Conference on Web Search and Data
  Mining}}. \bibinfo{pages}{586--595}.
\newblock


\bibitem[Sun et~al\mbox{.}(2025)]%
        {sun2025information}
\bibfield{author}{\bibinfo{person}{Zhouhao Sun}, \bibinfo{person}{Xiao Ding},
  \bibinfo{person}{Li Du}, \bibinfo{person}{Yunpeng Xu},
  \bibinfo{person}{Yixuan Ma}, \bibinfo{person}{Yang Zhao},
  \bibinfo{person}{Bing Qin}, {and} \bibinfo{person}{Ting Liu}.}
  \bibinfo{year}{2025}\natexlab{}.
\newblock \showarticletitle{Information Gain-Guided Causal Intervention for
  Autonomous Debiasing Large Language Models}.
\newblock \bibinfo{journal}{\emph{arXiv preprint arXiv:2504.12898}}
  (\bibinfo{year}{2025}).
\newblock


\bibitem[Tehrani et~al\mbox{.}(2024)]%
        {tehrani2024parenting}
\bibfield{author}{\bibinfo{person}{Hossein~Dabiriyan Tehrani},
  \bibinfo{person}{Sara Yamini}, {and} \bibinfo{person}{Alexander~T Vazsonyi}.}
  \bibinfo{year}{2024}\natexlab{}.
\newblock \showarticletitle{Parenting styles and Big Five personality traits
  among adolescents: A meta-analysis}.
\newblock \bibinfo{journal}{\emph{Personality and Individual Differences}}
  \bibinfo{volume}{216} (\bibinfo{year}{2024}), \bibinfo{pages}{112421}.
\newblock


\bibitem[Teng et~al\mbox{.}(2022)]%
        {teng2022understanding}
\bibfield{author}{\bibinfo{person}{Teng Teng}, \bibinfo{person}{Huifang Li},
  \bibinfo{person}{Yulin Fang}, {and} \bibinfo{person}{Lingzhi Shen}.}
  \bibinfo{year}{2022}\natexlab{}.
\newblock \showarticletitle{Understanding the differential effectiveness of
  marketer versus user-generated advertisements in closed social networking
  sites: An empirical study of WeChat}.
\newblock \bibinfo{journal}{\emph{Internet Research}} \bibinfo{volume}{32},
  \bibinfo{number}{6} (\bibinfo{year}{2022}), \bibinfo{pages}{1910--1929}.
\newblock


\bibitem[Van~Rooij et~al\mbox{.}(2024)]%
        {van2024reclaiming}
\bibfield{author}{\bibinfo{person}{Iris Van~Rooij}, \bibinfo{person}{Olivia
  Guest}, \bibinfo{person}{Federico Adolfi}, \bibinfo{person}{Ronald de Haan},
  \bibinfo{person}{Antonina Kolokolova}, {and} \bibinfo{person}{Patricia
  Rich}.} \bibinfo{year}{2024}\natexlab{}.
\newblock \showarticletitle{Reclaiming AI as a theoretical tool for cognitive
  science}.
\newblock \bibinfo{journal}{\emph{Computational Brain \& Behavior}}
  \bibinfo{volume}{7}, \bibinfo{number}{4} (\bibinfo{year}{2024}),
  \bibinfo{pages}{616--636}.
\newblock


\bibitem[Wang et~al\mbox{.}(2022)]%
        {wang2022hfenet}
\bibfield{author}{\bibinfo{person}{Di Wang}, \bibinfo{person}{Ronghao Yang},
  \bibinfo{person}{Hanhu Liu}, \bibinfo{person}{Haiqing He},
  \bibinfo{person}{Junxiang Tan}, \bibinfo{person}{Shaoda Li},
  \bibinfo{person}{Yichun Qiao}, \bibinfo{person}{Kangqi Tang}, {and}
  \bibinfo{person}{Xiao Wang}.} \bibinfo{year}{2022}\natexlab{}.
\newblock \showarticletitle{HFENet: hierarchical feature extraction network for
  accurate landcover classification}.
\newblock \bibinfo{journal}{\emph{Remote Sensing}} \bibinfo{volume}{14},
  \bibinfo{number}{17} (\bibinfo{year}{2022}), \bibinfo{pages}{4244}.
\newblock


\bibitem[Wang et~al\mbox{.}(2024)]%
        {wang2024multi}
\bibfield{author}{\bibinfo{person}{Hongyu Wang}, \bibinfo{person}{Dandan
  Zhang}, \bibinfo{person}{Jun Feng}, \bibinfo{person}{Lucia Cascone},
  \bibinfo{person}{Michele Nappi}, {and} \bibinfo{person}{Shaohua Wan}.}
  \bibinfo{year}{2024}\natexlab{}.
\newblock \showarticletitle{A multi-objective segmentation method for chest
  X-rays based on collaborative learning from multiple partially annotated
  datasets}.
\newblock \bibinfo{journal}{\emph{Information Fusion}}  \bibinfo{volume}{102}
  (\bibinfo{year}{2024}), \bibinfo{pages}{102016}.
\newblock


\bibitem[Wang et~al\mbox{.}(2023)]%
        {WangEtAl2023}
\bibfield{author}{\bibinfo{person}{Y. Wang}, \bibinfo{person}{D. Li},
  \bibinfo{person}{K. Funakoshi}, {and} \bibinfo{person}{M. Okumura}.}
  \bibinfo{year}{2023}\natexlab{}.
\newblock \showarticletitle{Emp: Emotion-guided multi-modal fusion and
  contrastive learning for personality traits recognition}. In
  \bibinfo{booktitle}{\emph{Proceedings of the 2023 ACM International
  Conference on Multimedia Retrieval}}. \bibinfo{pages}{243--252}.
\newblock


\bibitem[Wei et~al\mbox{.}(2022)]%
        {wei2022chain}
\bibfield{author}{\bibinfo{person}{Jason Wei}, \bibinfo{person}{Xuezhi Wang},
  \bibinfo{person}{Dale Schuurmans}, \bibinfo{person}{Maarten Bosma},
  \bibinfo{person}{Fei Xia}, \bibinfo{person}{Ed Chi}, \bibinfo{person}{Quoc~V
  Le}, \bibinfo{person}{Denny Zhou}, {et~al\mbox{.}}}
  \bibinfo{year}{2022}\natexlab{}.
\newblock \showarticletitle{Chain-of-thought prompting elicits reasoning in
  large language models}.
\newblock \bibinfo{journal}{\emph{Advances in neural information processing
  systems}}  \bibinfo{volume}{35} (\bibinfo{year}{2022}),
  \bibinfo{pages}{24824--24837}.
\newblock


\bibitem[William et~al\mbox{.}(2023)]%
        {william2023framework}
\bibfield{author}{\bibinfo{person}{P William}, \bibinfo{person}{N Yogeesh},
  \bibinfo{person}{Vishal~M Tidake}, \bibinfo{person}{Snehal~Sumit Gondkar},
  \bibinfo{person}{K Vengatesan}, {et~al\mbox{.}}}
  \bibinfo{year}{2023}\natexlab{}.
\newblock \showarticletitle{Framework for implementation of personality
  inventory model on natural language processing with personality traits
  analysis}. In \bibinfo{booktitle}{\emph{2023 International Conference on
  Intelligent Data Communication Technologies and Internet of Things
  (IDCIoT)}}. IEEE, \bibinfo{pages}{625--628}.
\newblock


\bibitem[Wu et~al\mbox{.}(2025a)]%
        {wu2025survey}
\bibfield{author}{\bibinfo{person}{Junchao Wu}, \bibinfo{person}{Shu Yang},
  \bibinfo{person}{Runzhe Zhan}, \bibinfo{person}{Yulin Yuan},
  \bibinfo{person}{Lidia~Sam Chao}, {and} \bibinfo{person}{Derek~Fai Wong}.}
  \bibinfo{year}{2025}\natexlab{a}.
\newblock \showarticletitle{A survey on LLM-generated text detection:
  Necessity, methods, and future directions}.
\newblock \bibinfo{journal}{\emph{Computational Linguistics}}
  (\bibinfo{year}{2025}), \bibinfo{pages}{1--66}.
\newblock


\bibitem[Wu et~al\mbox{.}(2025b)]%
        {wu2025interpreting}
\bibfield{author}{\bibinfo{person}{Xuansheng Wu}, \bibinfo{person}{Jiayi Yuan},
  \bibinfo{person}{Wenlin Yao}, \bibinfo{person}{Xiaoming Zhai}, {and}
  \bibinfo{person}{Ninghao Liu}.} \bibinfo{year}{2025}\natexlab{b}.
\newblock \showarticletitle{Interpreting and steering llms with mutual
  information-based explanations on sparse autoencoders}.
\newblock \bibinfo{journal}{\emph{arXiv preprint arXiv:2502.15576}}
  (\bibinfo{year}{2025}).
\newblock


\bibitem[Xue et~al\mbox{.}(2018)]%
        {xue2018deep}
\bibfield{author}{\bibinfo{person}{Di Xue}, \bibinfo{person}{Lifa Wu},
  \bibinfo{person}{Zheng Hong}, \bibinfo{person}{Shize Guo},
  \bibinfo{person}{Liang Gao}, \bibinfo{person}{Zhiyong Wu},
  \bibinfo{person}{Xiaofeng Zhong}, {and} \bibinfo{person}{Jianshan Sun}.}
  \bibinfo{year}{2018}\natexlab{}.
\newblock \showarticletitle{Deep learning-based personality recognition from
  text posts of online social networks}.
\newblock \bibinfo{journal}{\emph{Applied Intelligence}} \bibinfo{volume}{48},
  \bibinfo{number}{11} (\bibinfo{year}{2018}), \bibinfo{pages}{4232--4246}.
\newblock


\bibitem[Yang et~al\mbox{.}(2021a)]%
        {yang2021multi}
\bibfield{author}{\bibinfo{person}{Feifan Yang}, \bibinfo{person}{Xiaojun
  Quan}, \bibinfo{person}{Yunyi Yang}, {and} \bibinfo{person}{Jianxing Yu}.}
  \bibinfo{year}{2021}\natexlab{a}.
\newblock \showarticletitle{Multi-document transformer for personality
  detection}. In \bibinfo{booktitle}{\emph{Proceedings of the AAAI conference
  on artificial intelligence}}, Vol.~\bibinfo{volume}{35}.
  \bibinfo{pages}{14221--14229}.
\newblock


\bibitem[Yang et~al\mbox{.}(2021c)]%
        {yang2021learning}
\bibfield{author}{\bibinfo{person}{Feifan Yang}, \bibinfo{person}{Tao Yang},
  \bibinfo{person}{Xiaojun Quan}, {and} \bibinfo{person}{Qinliang Su}.}
  \bibinfo{year}{2021}\natexlab{c}.
\newblock \showarticletitle{Learning to answer psychological questionnaire for
  personality detection}. In \bibinfo{booktitle}{\emph{Findings of the
  Association for Computational Linguistics: EMNLP 2021}}.
  \bibinfo{pages}{1131--1142}.
\newblock


\bibitem[Yang et~al\mbox{.}(2023a)]%
        {yang2023orders}
\bibfield{author}{\bibinfo{person}{Tao Yang}, \bibinfo{person}{Jinghao Deng},
  \bibinfo{person}{Xiaojun Quan}, {and} \bibinfo{person}{Qifan Wang}.}
  \bibinfo{year}{2023}\natexlab{a}.
\newblock \showarticletitle{Orders are unwanted: dynamic deep graph
  convolutional network for personality detection}. In
  \bibinfo{booktitle}{\emph{Proceedings of the AAAI Conference on Artificial
  Intelligence}}, Vol.~\bibinfo{volume}{37}. \bibinfo{pages}{13896--13904}.
\newblock


\bibitem[Yang et~al\mbox{.}(2023b)]%
        {yang2023psycot}
\bibfield{author}{\bibinfo{person}{Tao Yang}, \bibinfo{person}{Tianyuan Shi},
  \bibinfo{person}{Fanqi Wan}, \bibinfo{person}{Xiaojun Quan},
  \bibinfo{person}{Qifan Wang}, \bibinfo{person}{Bingzhe Wu}, {and}
  \bibinfo{person}{Jiaxiang Wu}.} \bibinfo{year}{2023}\natexlab{b}.
\newblock \showarticletitle{PsyCoT: psychological questionnaire as powerful
  chain-of-thought for personality detection}.
\newblock \bibinfo{journal}{\emph{arXiv preprint arXiv:2310.20256}}
  (\bibinfo{year}{2023}).
\newblock


\bibitem[Yang et~al\mbox{.}(2021b)]%
        {yang2021psycholinguistic}
\bibfield{author}{\bibinfo{person}{Tao Yang}, \bibinfo{person}{Feifan Yang},
  \bibinfo{person}{Haolan Ouyang}, {and} \bibinfo{person}{Xiaojun Quan}.}
  \bibinfo{year}{2021}\natexlab{b}.
\newblock \showarticletitle{Psycholinguistic tripartite graph network for
  personality detection}.
\newblock \bibinfo{journal}{\emph{arXiv preprint arXiv:2106.04963}}
  (\bibinfo{year}{2021}).
\newblock


\bibitem[Yoon et~al\mbox{.}(2024)]%
        {yoon2024examining}
\bibfield{author}{\bibinfo{person}{Hee~Jun Yoon}, \bibinfo{person}{Brent~W
  Roberts}, \bibinfo{person}{Madison~N Sewell}, \bibinfo{person}{Christopher~M
  Napolitano}, \bibinfo{person}{Christopher~J Soto}, \bibinfo{person}{Dana
  Murano}, {and} \bibinfo{person}{Alex Casillas}.}
  \bibinfo{year}{2024}\natexlab{}.
\newblock \showarticletitle{Examining SEB skills’ incremental validity over
  personality traits in predicting academic achievement}.
\newblock \bibinfo{journal}{\emph{Plos one}} \bibinfo{volume}{19},
  \bibinfo{number}{1} (\bibinfo{year}{2024}), \bibinfo{pages}{e0296484}.
\newblock


\bibitem[Zelikman et~al\mbox{.}(2024)]%
        {zelikman2024star}
\bibfield{author}{\bibinfo{person}{Eric Zelikman}, \bibinfo{person}{YH Wu},
  \bibinfo{person}{Jesse Mu}, {and} \bibinfo{person}{Noah~D Goodman}.}
  \bibinfo{year}{2024}\natexlab{}.
\newblock \showarticletitle{STaR: Self-taught reasoner bootstrapping reasoning
  with reasoning}. In \bibinfo{booktitle}{\emph{Proc. the 36th International
  Conference on Neural Information Processing Systems}},
  Vol.~\bibinfo{volume}{1126}.
\newblock


\bibitem[Zhu et~al\mbox{.}(2024a)]%
        {zhu2024enhancing}
\bibfield{author}{\bibinfo{person}{Haohao Zhu}, \bibinfo{person}{Xiaokun
  Zhang}, \bibinfo{person}{Junyu Lu}, \bibinfo{person}{Youlin Wu},
  \bibinfo{person}{Zewen Bai}, \bibinfo{person}{Changrong Min},
  \bibinfo{person}{Liang Yang}, \bibinfo{person}{Bo Xu},
  \bibinfo{person}{Dongyu Zhang}, {and} \bibinfo{person}{Hongfei Lin}.}
  \bibinfo{year}{2024}\natexlab{a}.
\newblock \showarticletitle{Enhancing Textual Personality Detection toward
  Social Media: Integrating Long-term and Short-term Perspectives}.
\newblock \bibinfo{journal}{\emph{arXiv preprint arXiv:2404.15067}}
  (\bibinfo{year}{2024}).
\newblock


\bibitem[Zhu et~al\mbox{.}(2024b)]%
        {zhu2024integrating}
\bibfield{author}{\bibinfo{person}{Haohao Zhu}, \bibinfo{person}{Xiaokun
  Zhang}, \bibinfo{person}{Junyu Lu}, \bibinfo{person}{Liang Yang}, {and}
  \bibinfo{person}{Hongfei Lin}.} \bibinfo{year}{2024}\natexlab{b}.
\newblock \showarticletitle{Integrating multi-view analysis: Multi-view
  mixture-of-expert for textual personality detection}. In
  \bibinfo{booktitle}{\emph{CCF International Conference on Natural Language
  Processing and Chinese Computing}}. Springer, \bibinfo{pages}{359--371}.
\newblock


\bibitem[Zhu et~al\mbox{.}(2025)]%
        {zhu2025investigating}
\bibfield{author}{\bibinfo{person}{Jianfeng Zhu}, \bibinfo{person}{Ruoming
  Jin}, {and} \bibinfo{person}{Karin~G Coifman}.}
  \bibinfo{year}{2025}\natexlab{}.
\newblock \showarticletitle{Investigating Large Language Models in Inferring
  Personality Traits from User Conversations}.
\newblock \bibinfo{journal}{\emph{arXiv preprint arXiv:2501.07532}}
  (\bibinfo{year}{2025}).
\newblock


\bibitem[Zhu et~al\mbox{.}(2022a)]%
        {zhu2022contrastive}
\bibfield{author}{\bibinfo{person}{Yangfu Zhu}, \bibinfo{person}{Linmei Hu},
  \bibinfo{person}{Xinkai Ge}, \bibinfo{person}{Wanrong Peng}, {and}
  \bibinfo{person}{Bin Wu}.} \bibinfo{year}{2022}\natexlab{a}.
\newblock \showarticletitle{Contrastive Graph Transformer Network for
  Personality Detection.}. In \bibinfo{booktitle}{\emph{IJCAI}}.
  \bibinfo{pages}{4559--4565}.
\newblock


\bibitem[Zhu et~al\mbox{.}(2022b)]%
        {zhu2022lexical}
\bibfield{author}{\bibinfo{person}{Yangfu Zhu}, \bibinfo{person}{Linmei Hu},
  \bibinfo{person}{Nianwen Ning}, \bibinfo{person}{Wei Zhang}, {and}
  \bibinfo{person}{Bin Wu}.} \bibinfo{year}{2022}\natexlab{b}.
\newblock \showarticletitle{A lexical psycholinguistic knowledge-guided graph
  neural network for interpretable personality detection}.
\newblock \bibinfo{journal}{\emph{Knowledge-Based Systems}}
  \bibinfo{volume}{249} (\bibinfo{year}{2022}), \bibinfo{pages}{108952}.
\newblock


\bibitem[Zubi{\'c} et~al\mbox{.}(2024)]%
        {zubic2024limits}
\bibfield{author}{\bibinfo{person}{Nikola Zubi{\'c}}, \bibinfo{person}{Federico
  Sold{\'a}}, \bibinfo{person}{Aurelio Sulser}, {and} \bibinfo{person}{Davide
  Scaramuzza}.} \bibinfo{year}{2024}\natexlab{}.
\newblock \showarticletitle{Limits of Deep Learning: Sequence Modeling through
  the Lens of Complexity Theory}.
\newblock \bibinfo{journal}{\emph{arXiv preprint arXiv:2405.16674}}
  (\bibinfo{year}{2024}).
\newblock


\end{thebibliography}

\end{document}